\documentclass[journal]{IEEEtran}
\usepackage{amsmath,amssymb}
\usepackage{booktabs}
\usepackage{graphicx}
\usepackage{multirow}
\usepackage[hidelinks]{hyperref}

\newcommand{\pw}{our earlier work~\cite{paper1}}
\newcommand{\Pw}{Our earlier work~\cite{paper1}}
\newcommand{\pwof}{our earlier work}
\newcommand{\repourl}{\url{https://github.com/maxsegan/moba-draft-policies}}
\newcommand{\osfurl}{\url{https://osf.io/3cxvd}}

\begin{document}

\title{Drafting for the Players:\\ Player-Aware Values and Proposals for MOBA Draft Search}
\author{Max~Segan, Ernest~Pysher, and Hod~Lipson%
\thanks{M. Segan is with the Department of Computer Science and H. Lipson with the Department of Mechanical Engineering, Columbia University (e-mail: m.segan@columbia.edu; hod.lipson@columbia.edu). E. Pysher is an independent researcher (e-mail: ernestpysheriv@gmail.com).}}

\maketitle

\begin{abstract}
A draft model for a multiplayer online battle arena game is a learned value function that scores a draft by its chance of winning, and a search drafter picks heroes by optimizing it. On 84,658 ranked Heroes of the Storm games from a later, untouched patch, adding the ten players raises accuracy from 57.2\% to 62.7\%. We score drafters by the players' later real results on the heroes each drafter gives them. Search over a hero-only value shows no detected gain over one-step greedy drafting, and adding the players to the value alone loses 3.3 points to player-aware greedy, because the candidate heroes still come from population behavior. When both the value and the candidates know the players, our drafter beats player-aware greedy by 2.2 points and the players' own drafts by 2.8 on those later results. The gain lies in the learned player-aware prior: picking straight from it, with no search at play time, is not detectably behind the full drafter, while a flat prior over greedy's candidates loses 3.6 points. The drafter stays on well-covered compositions, with 1.3\% of its teams below a support floor against 6.3\% of human teams.
\end{abstract}


\section{Introduction}
\label{sec:intro}

A multiplayer online battle arena (MOBA) game has two stages. First, two teams of five take turns banning and picking heroes. Then they play the match. A drafting system needs a value function, a model that scores a draft by its chance of winning, and a way to optimize it. Work on drafting learns that value function from logged games and searches or recommends against it~\cite{chen2018art,summerville2016draft,gourdeau2021discriminative,chen2021juewudraft}. \Pw{} built such a search drafter for Heroes of the Storm (HotS) and studied when search over a learned value function can be trusted to leave the data. From the heroes, the map and the skill tier alone, its value function calls 57.2\% of the spring 2026 ranked games we test on here. We report accuracy as an \emph{edge over a coin flip}, accuracy minus 50, so a 57.2\% model has a 7.2-point edge.

That value function, like most draft models, sees only the heroes. A ranked game is played by ten particular people, and who plays a hero matters: experienced players do about 7 points worse than the draft predicts on a hero they have never played. A search drafter that optimizes a hero-only value ignores which player gets which hero.

\textbf{The value function.} On 84,658 games from a later patch that no step of the work had seen, adding each player's history with their hero raises accuracy from 57.2\% to 62.7\%, an edge of 12.7 points against 7.2. The model separates how strong a hero is for everyone from how well each player does on it, weights each record by how much of it carries forward, and reads the scoreboard as well as the result.

\textbf{Search.} We compare search with one-step greedy drafting on the same value function, measured by the players' later real results on the heroes each drafter gives them. With a population value, search shows no detected gain over greedy on this measure ($-$1.3 points, 95\% CI $-$4.7 to 2.1). Adding the players to the value alone makes search lose to player-aware greedy ($-$3.3, $-$5.9 to $-$0.8): its candidate heroes still come from population behavior, so it keeps evaluating players on heroes they rarely play. Making the candidates player-aware as well gains 4.5 points (2.1 to 7.0) at the same value. Our final drafter beats player-aware greedy by 2.2 points (0.4 to 3.9) and the players' own drafts by 2.8 (1.0 to 4.6). Picking straight from its learned prior, with no search at play time, it beats player-aware greedy by 2.2 points (0.3 to 4.0) and is not detectably behind the full drafter (0.4, $-$1.2 to 2.0). With a flat prior over greedy's 30 candidates and greedy's bans it loses 3.6 points (1.7 to 5.5). The gain comes from how the learned prior ranks the heroes, not from decision-time lookahead or budget. A natural experiment shows what a hero is worth to its player: when a player's best hero is banned or taken before they pick (16\% of slots), they do 1.99 points worse than the draft alone predicts, against 1.45 predicted by the player model.

\textbf{Support.} \Pw{} found that search leaves the data in composition space. With players there is a second axis: the player on each hero. On composition, 1.3\% of our drafter's teams fall below a support floor, against 6.3\% for humans. On the player axis, it places players on heroes they barely know about as often as the humans do under the same assignment rule (9.9\% against 9.4\% with fewer than 5 earlier games), though more of them on heroes they have never played (3.8\% against 2.3\%).

\textbf{Parties and upkeep.} Ranked players queue alone or in small parties, so we also draft for part of a team. In simulation, a solo player gets a fifth of what a full team gets, more with the bans, and does best by locking in last. Both halves of the value function age: hero strength keeps a correlation of 0.85 over a year and skill on frequently played heroes 0.74.

Our contributions:
\begin{enumerate}
\item A player-aware value function that separates hero strength from player skill and weights each record by measured evidence of how much of it carries forward (Section~\ref{sec:players}).
\item A ladder of search drafters over population and player-aware values and proposals, measured against greedy drafting and the players' own drafts by later real results, which shows that player knowledge pays off only when it reaches the proposals, and controls showing that the gain over greedy comes from the learned prior's ranking: picking from that prior with no search is not detectably behind the full drafter, and decision-time lookahead and budget add no detected gain (Section~\ref{sec:drafting}).
\item Support on two axes, composition and player, with a floor and a calibration of the player term that keep the drafter on support on both axes at no measured cost (Section~\ref{sec:support}), and an account of where the value function is still wrong (Section~\ref{sec:wrong}).
\item A drafter for parties of one to five (Section~\ref{sec:party}) and an upkeep policy from measured rates of change on both sides (Section~\ref{sec:upkeep}).
\end{enumerate}

\section{Related Work}
\label{sec:related}

\textit{Draft prediction and drafting.} Draft models estimate win probability from the heroes and search or recommend against it~\cite{chen2018art,chen2021juewudraft}, or imitate expert picks~\cite{summerville2016draft,gourdeau2021discriminative}. OpenAI Five drafted by search over a restricted hero pool~\cite{berner2019dota}. \Pw{} measured how far search over a hero-level value leaves the data, found the value no more wrong on rare compositions than common ones, and showed that a support floor costs little when the prior is trained under it. Draft-only outcome prediction for Dota 2 compares standard learners~\cite{semenov2016performance}, and association rules mined from line-ups recommend heroes~\cite{hanke2017recommender}. All of these treat a draft as heroes, a map and a tier. Win prediction from in-match data uses the players' performance as the game unfolds~\cite{hodge2021win}.

\textit{Players.} Rating systems such as Elo, Glicko and TrueSkill~\cite{elo1978rating,glickman1999parameter,herbrich2006trueskill} give each player one number, which is what matchmakers balance. TrueSkill 2 adds experience and per-game scores to a rating~\cite{minka2018trueskill2}, still one number per player. Team performance in League of Legends depends on more than individual ratings~\cite{kim2017team}, including how well each player's role experience fits the team~\cite{kim2016proficiency}, and individual performance in team games varies with play history~\cite{sapienza2018individual}. Do et al.~\cite{do2021player} predict League of Legends outcomes from player-champion experience features after the draft. DraftRec~\cite{draftrec} is the closest drafting system to ours: a transformer over each player's recent games that recommends picks and predicts the winner. We use it as a baseline and opponent model. We go beyond these in three ways: we measure how much of a player's record on a hero is real and how fast it changes, we search drafts with the players' values, and we judge drafters with judges fit apart from them, including one whose player model shares no fitting games with the drafters', and with an observed-outcome check.

\textit{Methods.} Shrinking short records toward a population value is empirical Bayes~\cite{efron1975data}, familiar from batting averages~\cite{brown2008inseason}. Our model's slow drift is a state-space model~\cite{kalman1960new}, and borrowing across heroes resembles matrix factorization with time effects~\cite{koren2009matrix,koren2009temporal}. Correlations between noisy measurements are corrected for noise as in Spearman~\cite{spearman1904proof}. Drift and its maintenance are surveyed in~\cite{gama2014survey,lu2019drift,quinonero2009dataset,sculley2015debt}. The drafters use Monte Carlo tree search~\cite{browne2012survey} with a learned prior in the AlphaZero style~\cite{silver2018general,rosin2011multi}, and tournaments are summarized with Bradley-Terry strengths~\cite{bradley1952rank}.

\section{Data and Evaluation}
\label{sec:data}

\subsection{Games, players and windows}

We use Storm League, the ranked draft mode of HotS, whose skill tiers run from Bronze through Master. Our development corpus holds every Storm League game we have from December 2021 through September 27, 2026, 22.4 million player slots with each player's hero, scoreboard and result. Our models use the 90 heroes of the pool before patch 2.57. Games come from the Americas, Europe and Korea; China is not in the data. For players in these games we also pulled full match histories from Heroes Profile, the public database our replays come from, which gives a history store of 59.7 million player games.

Table~\ref{tab:windows} lists the windows. The player model is fit on games up to the draft model's training cutoff, which falls during February 10, 2026. Its combining stages, from the trust curve to the game combiner, are fit on a calibration window of the games after that cutoff, so no game is in both. The \emph{spring test set} and the \emph{summer test set} measure it. The \emph{post-freeze set}, games of patch 2.57 played from September 28, 2026, the day after our data cutoff and pulled once for this score, is the untouched confirmation: no step of the work read them before the model was frozen. The drafters' outcome data end on May 22, 2026, before the tournament lobbies. The opponent model's three DraftRec coefficients and the player-aware prior's search targets are fit on real actions from 3,000 tournament-window lobbies that share no game with the tournament, the 300 tune lobbies or the party study; no outcome enters either fit.

\begin{table}[t]
\centering
\caption{Windows. $^\dagger$The fitting window ends at the draft model's training cutoff, partway through February 10, 2026; the calibration window holds only games after that cutoff, so the two share no game.}
\label{tab:windows}
\setlength{\tabcolsep}{3pt}
\begin{tabular}{llr}
\toprule
Window & Dates (2026) & Games \\
\midrule
Calibration (combining weights) & Feb 10$^\dagger$ to Mar 28 & 70,974 \\
Spring test & Mar 29 to May 22 & 72,474 \\
Summer test & May 23 to Sep 27 & 286,016 \\
Post-freeze (patch 2.57) & Sep 28 to Oct 26 & 84,658 \\
\midrule
Tournament lobbies & May 23 to Jun 30 & 300 \\
Party-study lobbies & May 23 to Jun 30 & 1,175 \\
Judge window & Jul 1 to Sep 11 & 144,414 \\
\bottomrule
\end{tabular}
\end{table}

\textit{Provenance and leakage.} Every input is dated before the game it predicts. Player records and scoreboard profiles exclude the predicted day, and ratings use the last value before the game starts. Match histories are dated by game end, so a record can include a game played before the predicted day but uploaded after it. Spring and summer games supported model selection, so their gains are development estimates. The post-freeze set touched no step of model selection: we scored it once with every weight frozen and player records updating daily. Of its 95,973 games, 7,341 with the new hero and 3,974 with a missing player row drop out, leaving 84,658 (88\%). In the tournament, the judge's and the drafters' player models are fit on disjoint halves of the players but share game outcomes; Section~\ref{sec:robust} refits the judge without them. The supplement lists what each component is fit on and how it updates.

\subsection{The draft alone}

The draft-only model has the architecture and 283 inputs of the value network of \pw{} (heroes, map, tier, and per-tier hero, hero-map and hero-pair statistics), retrained here on games before February 10, 2026, with statistics from earlier builds only, recomputed after every build (the random-split model of \pw{} is not used). It is right on 57.19\% of spring games, a 7.19-point edge, and 56.94\% of summer games. The draft explains 3.4\% of the variance of one game's result.

\section{Adding the Players}
\label{sec:players}

Adding the players raises accuracy from 57.22\% to 62.69\% on the post-freeze set and from 57.19\% to 63.12\% on the spring test set (Table~\ref{tab:headline}).

\subsection{Separating the hero from the player}

The draft-only model already prices how strong a hero is for everyone. The player model measures only how a player does on that hero compared with what the draft-only model expected. For games we know only from Heroes Profile match histories, where we have no draft-only prediction, the baseline is the hero's population win rate in its quarter (quarters after fitting reuse the last fitted one), which carries no information about the player. We call the running average of these differences a player's \emph{record} on the hero, in points of win probability. To predict a game, the model sums each team's player estimates and combines the difference between the teams with the draft-only prediction $W$:
\begin{multline}
\mathrm{logit}\,P(A) = c_0 + c_1\,\mathrm{logit}\,W \\ + c_2\Big(\sum_{p\in A} s_p - \sum_{p\in B} s_p\Big) + c_3(a_A - a_B),
\end{multline}
where $s_p$ is player $p$'s estimate on their hero $h$ (as a probability fraction) and $a_A$ sums the team's slot-level rating terms. At its core $s_p = m_0 + w\,(\bar y_{ph} - m_0)$: $\bar y_{ph}$ is the player's precision-weighted mean of $y - W$ on $h$, net of an experience offset, discounted each day by a random-walk drift, $m_0$ a prior from the player's other records and scoreboard (below), and $w = P/(P + 1/v_0)$ the weight a Gaussian model gives a record of precision $P$ against a prior of variance $v_0$. The full model rescales these terms by smooth functions of games on the hero, its share of the player's games and time since last played, and adds the scoreboard and short-term terms below; the $c$ are fit on the calibration window (full specification in the supplement).

Matchmaking equalizes ratings, but a rating averages over all of a player's heroes, and the matchmaker cannot see which hero a player will pick. On their best hero a player is better than their rating says, and on a hero they rarely play they are worse. Each player's stable hero-specific effect is about 0.45\% of one game's variance (a split-half covariance between periods three months apart). Ten such effects, if independent of each other and of the draft, would sum to 4.5\%, comparable to the 3.4\% variance of the draft-only prediction; the two are estimated differently, so the comparison is rough. Players with 300 or more earlier games keep about the same share (0.51\%), so it does not come from new accounts still climbing.

\subsection{How much to trust a record}

One game is close to a coin flip, so a short record is mostly luck. For every game in the spring and summer sets we took each player's record on their hero up to the day before and measured how much of it showed up in that game. A record of 30 to 59 games carries 24\% into the next game, and one of 800 or more carries 81\% (supplement). Most of the shortfall is noise: simulated players whose true skill never changes, with the real games and dates, return the noise-only share within 0.03 in every bin, and the real data sit at 0.79 to 0.85 of it for records of 30 to 799 games and at 1.02 for 800 or more. Adding both teams' ratings at game time moves these shares by 0.02 or less, so matchmaking does not hide them. Edges on a hero that is over half of a player's games persist less (0.56 to 0.66) than on heroes under 15\% of their games (0.91 to 0.96).

A single rule, weight = games / (games + 180), follows the trend. It measures what carries into future games, drift included, so it reaches half later than the within-period consistency of Section~\ref{sec:scoreboard} (74 games). The full model learns this curve, the trust curve, instead (Section~\ref{sec:players}-A). The result is close to calibrated where most records are. Regressing realized results on predictions by games on the hero, the slope is 1.05 at 0 games, 0.95 at 30, 1.00 at 100 and 0.89 at 300, against 1.31 to 0.76 for the base model.

\subsection{Discounting older records}

A record can go stale. The model lets each player's level on each hero drift as a random walk, tracked with a Kalman-style filter~\cite{kalman1960new} over each player's whole history. Only the drift's size is chosen on the calibration window, and values from 1.5 to 3 points a year score almost the same. The one-year drift standard deviation is about 2 points of win probability. A 100-game record keeps half its precision through about 6 years without play. The scoreboard profile drifts faster: a 100-game profile keeps half its precision for about 10 days (2 days to 5 months, depending on the statistic). A short-term layer reads each player's recent form (45-day half-life; a recent run above expectation is partly given back) and their results on heroes they recently picked up, which carries most of its value.

\subsection{Borrowing from related play}

Many slots are players on heroes they have barely played: the median player in a test game has 17 games on their hero. Three things fill the gap. \textit{Experience}: how many games a player has on each hero is strong evidence before any results. Veterans do about 7 points worse than the draft predicts on heroes they have never played and 2 to 3 points better on heroes they play a lot. \textit{Overall level}: each player-hero record is shrunk toward a prior built from the player's overall level, their records on heroes with similar traits and their scoreboard. \textit{Learned hero traits}: the model learns a few unlabeled traits per hero from the whole player base, so that heroes sharing traits are played well by the same people, and blends a thin record with what the player's records on similar heroes imply. With a hero's own games hidden, borrowing predicts 30 to 38\% of the real variation in a player's skill on it. Its value is on heroes a player has barely played; aggregate accuracy barely moves.

\subsection{How a player played, as well as whether they won}
\label{sec:scoreboard}

A win or a loss says little about one player, because nine other people shaped it. A stable effect of that player on that hero, counting their overall level and their edge on the hero together, is 1.33\% of the variation in one result beyond the draft within a period, so a win-loss record needs about 74 games before half of it is real. (Across periods three months apart, as in Section~\ref{sec:players}-A, the stable hero edge alone is 0.45\%.) The scoreboard is far more consistent: half of a record is real after 1.5 games for protection given to allies, 3.4 for hero damage and 14 for kills (split-half reliability on 9.4 million fitting-window rows; supplement). Stats a player controls alone repeat from game to game; stats that depend on the other nine, such as kills, swing with who else is in the game.

Much of what repeats is play style, so the model learns which consistent traits go with winning and weights those. It uses the scoreboard twice. One path runs through the learned hero traits. The other is direct: a ridge regression on 42 running averages of each player's normalized numbers and team shares, on the hero and across heroes, with weights that vary by hero role. Among the full model's additions, the direct path is the largest gain in log loss.

\textit{One scale for every player.} Raw scoreboards are not comparable. Uther heals less than other healers because part of his value is armor, long games inflate totals, and winners' numbers look better because they won. Each number becomes a compressed per-minute rate, standardized against games on the same hero and map, game-length band and result, so it adds information the win record does not carry. It is capped, adjusted per hero and quarter for balance patches, and joined by the player's share of team totals (supplement). Reasonable alternatives move accuracy by at most about 0.2 points.

\subsection{How the edge builds}

\begin{figure}[t]
\centering
\includegraphics[width=\columnwidth]{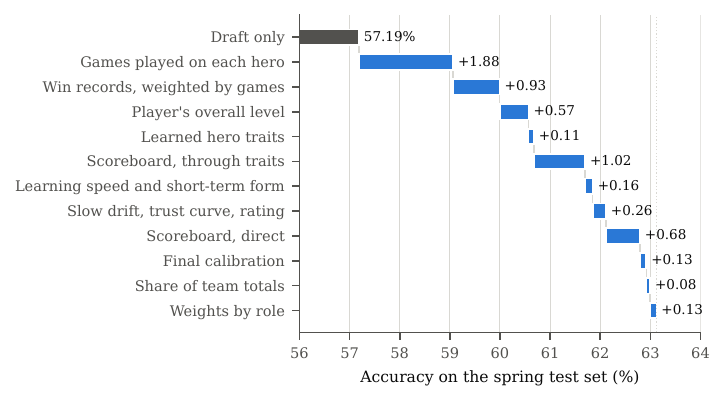}
\caption{What each player-model step adds to accuracy on the spring test set (72,474 games). Each bar is scored on the same games.}
\label{fig:ladder}
\end{figure}

Fig.~\ref{fig:ladder} adds the pieces one at a time on the spring test set, each step scored on the same games with its combining weights refit on the calibration window. The single biggest step is knowing how much each player has played each hero (+1.88 points, standard error (SE) 0.17), before any results. Win records weighted by their games add 0.93 accuracy points, and the player's overall level adds 0.57. Learned hero traits add 0.11 points, the scoreboard through the hero traits adds 1.02, and the short-term layer adds 0.16, reaching 61.85\%. The full model adds slow drift, the learned trust curve, a game-level rating term, the direct scoreboard channel, calibration, team shares and role weights, 1.27 points in all, reaching 63.12\%; the direct channel is the largest of these at 4.8 thousandths of log loss.

\subsection{Result}

Table~\ref{tab:headline} gives the full model against the draft-only model and two simpler player models: a win-only model (through learned hero traits in Fig.~\ref{fig:ladder}) and a base model (through the short-term layer). On the spring test set the edge grows from 7.19 to 13.12 points, 82\% larger; the gain of 5.93 points has a week-block 95\% confidence interval (CI) of 5.62 to 6.29. On the summer test set it grows from 6.94 to 12.85, 85\% larger (gain 5.91, CI 5.73 to 6.08). Ratings alone do not explain the gain: adding the players' ratings at game time, overall or on their heroes, to the draft-only model raises accuracy by at most 0.6 points and improves log loss by at most 2.8 thousandths, against 36.53 for the full model. The closest prior system predicts far less: on the spring games DraftRec's win head can score, it reaches 60.5\% accuracy, and its log-loss gains over draft only are 15.1 thousandths in spring and 15.6 in summer, against 36.0 and 37.3 for the full model on the same games (supplement).

\begin{table}[t]
\centering
\caption{Accuracy (\%) of the draft-only model and three player models, and their log-loss gain over the draft-only model (thousandths). Post-freeze: patch 2.57, scored once with no refit.}
\label{tab:headline}
\setlength{\tabcolsep}{3.5pt}
\begin{tabular}{lccc}
\toprule
Model & Spring & Summer & Post-freeze \\
\midrule
Draft only & 57.19 & 56.94 & 57.22 \\
Win records only & 60.68 & 60.33 & 60.20 \\
Base player model & 61.85 & 61.70 & 61.59 \\
Full player model & \textbf{63.12} & \textbf{62.85} & \textbf{62.69} \\
\midrule
\multicolumn{4}{l}{\textit{Log-loss gain over draft only}} \\
Win records only & 19.22 & 18.49 & 15.87 \\
Base player model & 29.22 & 28.31 & 25.34 \\
Full player model & \textbf{36.53} & \textbf{37.38} & \textbf{32.95} \\
\midrule
Games & 72,474 & 286,016 & 84,658 \\
\bottomrule
\end{tabular}
\end{table}

\textit{The untouched confirmation.} The post-freeze set is the 84,658 games of patch 2.57 in our database whose ten heroes are all among the model's 90 and whose ten players all have rows in our player table; games with the new hero, Xal'atath, drop out. The draft alone is right 57.22\% of the time, the base model 61.59\% and the full model 62.69\%, so the players add 5.47 points (CI 5.12 to 5.79). The full model's log-loss gain over the base model is 7.61 thousandths, against 9.07 on the summer test set. Four weeks give few week blocks, so we give both intervals: 6.82 to 8.07 by week and 6.36 to 8.87 by 3-day blocks. The calibration slope is 0.956 overall; for players with 0 to 3 earlier games on their hero it is 0.84 to 0.87.

\textit{Where it helps most.} The gain holds in every tier, from 4.8 points in Bronze to 7.0 in Diamond on the spring set, and in every region (supplement). On the base model, the 5\% of slots with the largest personal estimates in absolute value carry 38\% of the gain, and 65\% of those are negative: players on heroes they have barely played, who do 8.4 points worse than the draft expects. For scouting, this suggests most value in knowing when an opponent is off their usual pool; bans aimed at opponents' best heroes are the one design-study part that helped on both lobby sets.

\section{Searching with the Players}
\label{sec:drafting}

We test drafters as in \pw: every drafter drafts for both sides of real lobbies, against every other. Each draft is scored two ways. The primary measure uses no model of the players: it reads the players' later real results on the heroes a drafter gave them. The secondary measure is a judge, a model that sees the players through a separate half of the player base.

\subsection{The drafters}

Every drafter picks and bans for one team in the Storm League order (bans 0, 1, 0, 1; picks 0, 1, 1, 0, 0; bans 1, 0; picks 1, 1, 0, 0, 1). A team may pick only heroes that at least one of its players has played before the lobby day; if none is free, any free hero is allowed. A team's personal value uses its best assignment of heroes to players over all 120 permutations, so the real pick order is never used. Every search drafter plays under the structural mask of \pwof{}, which keeps a hero only if the team can still field a healer, a frontline hero and a ranged damage dealer without stacking a role.

A search drafter has two parts that can know the players or not. The \emph{value} scores a finished draft. The \emph{proposals} are the prior that decides which heroes the search considers and how often.

\begin{itemize}
\item \textbf{Population search}, from \pw. Monte Carlo tree search at decision time with 400 simulations per move, a prior trained by MCTS against imitation opponents, and the draft-only value. At opponent turns the search samples moves from the imitation drafter (below).
\item \textbf{Player-aware value, population proposals.} The same search, with the players added to the value: $V = \sigma(b_1\,\mathrm{logit}\,W + b_2(S_\text{ours} - S_\text{theirs}) + b_3(O_\text{ours} - O_\text{theirs}))$, where $W$ is the draft-only win probability, $S$ a team's summed player estimates on its heroes, and $O$ its count of players on a hero outside the roles they usually play; $b_1$ to $b_3$ are fit on held-out games.
\item \textbf{Player-aware value and proposals.} The same value with a prior that also sees the players. The prior logits are $\alpha \log p_F + \log \mathrm{mean}_{\text{team}} \exp g(f(p,h))$, where $p_F$ is the population prior and $g$ a two-layer network on six features of player $p$ on hero $h$ (the estimate, games played, whether never played, share of the player's games, and two role terms). It is distilled from a player-aware search at 1,600 simulations on 30,000 real pick states from non-tournament lobbies; the fitted $\alpha$ is 0.16.
\item \textbf{Selected design.} The previous drafter plus three parts that use all ten players: bans aimed at the opponents' best heroes, a model of each opponent's picks at chance nodes, and a bias toward heroes that improve our team's best assignment. A rule fixed in advance chose it on 300 separate tune lobbies.
\item \textbf{Final drafter.} The selected design with a composition support floor and a calibrated player term (Section~\ref{sec:support}).
\item \textbf{Greedy}, player-aware and population. At each pick it scores the top 15 heroes by the imitation model and the top 15 by the team's personal values, with 4 rollouts each, and takes the best.
\item \textbf{DraftRec}~\cite{draftrec}, rebuilt from its paper and released code and trained on 862,604 of our drafts through May 22, 2026, as its pick head and as its pick-and-win rule.
\item \textbf{Imitation drafter.} The five behavioral-cloning models of \pwof{}, which pick what real players most often pick.
\item \textbf{The players' own drafts.} The lobby's real picks and bans, replayed while legal, with the imitation drafter's choice when a real action is no longer legal.
\end{itemize}

\subsection{Lobbies and measures}

The 300 tournament lobbies are real games from May 23 to June 30, 2026, after every drafter's outcome data. In each, all ten players have 50 or more earlier games and 10 or more games in the judge window, so the tournament measures drafting for established, active players. Every pair of drafters plays every lobby from both sides, and the drafters read each player's estimates for all 90 heroes as of the day before the lobby.

\textit{The paired record.} For each drafter we assign its heroes to the players by their earlier play counts, a rule that reads no outcomes. We then take each player's later real record on that hero: their results in judge-window games on it (July 1 to September 11), minus the draft-only expectation. Two drafters are compared on the players covered under both. The paired record uses no model of the players. It reads heroes players chose themselves in later games, not heroes assigned to them, and covers only players with later games on both drafters' heroes (29 to 64\% here, mostly established players). It therefore tests whether the player edges a drafter builds on persist; it is not a win rate for following the drafter. An observational backtest on real games checks the direction (Section~\ref{sec:robust}).

\textit{Judges.} The population judges are four hero-level models from \pwof{}, fit only on games no drafter saw (later ranked uploads, including the lobbies' own games, and Quick Match games), each corrected for its under-penalty of teams missing a required role. The player-aware judge (J) combines their consensus with a personal term whose player model is fit on the other half of the players, with evidence from the judge window. The population judges see only the heroes. A drafter's judged score against another is its drafts' mean judged win probability against the other's, over the 300 lobbies and both seats. The judge's personal term runs on a smaller scale than the drafters' (0.55 per unit), and a judge refit on spring games alone shares no fitting games with the drafters (Section~\ref{sec:robust}). The population judges' fitting games include the real games of 299 of the 300 lobbies, which no drafter saw; they are 299 of 260,539 fitting games, and a Quick Match judge that contains none of them moves every population-judge comparison by at most 2.4 points and reverses none (supplement).

\subsection{The search ladder}

\begin{table*}[t]
\centering
\caption{The search ladder on the 300 tournament lobbies. Paired record: later real results minus draft-only expectation, points per covered player, with 95\% CI clustered by lobby and by player (rows held to the root or without search: bootstrap over lobbies). Coverage: share of players covered under both drafters. J and population judges: mean judged win probability (\%) of the first drafter.}
\label{tab:ladder}
\setlength{\tabcolsep}{4pt}
\begin{tabular}{llcccc}
\toprule
Search & Compared with & Paired record & Coverage & J & Population judges \\
\midrule
Population value, population proposals & population greedy & $-$1.3 ($-$4.7, 2.1) & 29\% & 52.4 & 56.7 \\
Player-aware value, population proposals & player-aware greedy & $-$3.3 ($-$5.9, $-$0.8) & 34\% & 47.8 & 57.5 \\
Player-aware value and proposals & player-aware greedy & +1.1 ($-$0.5, 2.7) & 55\% & 55.2 & 53.8 \\
Selected design & player-aware greedy & +3.1 (1.1, 5.1) & 54\% & 57.5 & 53.4 \\
Final drafter & player-aware greedy & \textbf{+2.2} (0.4, 3.9) & 53\% & 56.9 & 53.3 \\
Final drafter, held to the root & player-aware greedy & \textbf{+3.3} (1.3, 5.3) & 53\% & 56.4 & 53.4 \\
Final drafter's prior, no search & player-aware greedy & \textbf{+2.2} (0.3, 4.0) & 55\% & 54.1 & 50.8 \\
\midrule
Final drafter & final drafter, held to the root & +0.3 ($-$1.3, 1.8) & 58\% & 50.2 & 49.9 \\
Final drafter & its prior, no search & +0.4 ($-$1.2, 2.0) & 57\% & 52.2 & 52.3 \\
Held to the root & same, with greedy's budget & $-$0.3 ($-$1.7, 1.1) & 60\% & 50.5 & 50.7 \\
Held to the root & same, flat prior on greedy's candidates and bans & \textbf{+3.6} (1.7, 5.5) & 55\% & 55.6 & 52.4 \\
Player-aware value and proposals & player-aware value, population proposals & \textbf{+4.5} (2.1, 7.0) & 38\% & 57.8 & 44.6 \\
\bottomrule
\end{tabular}
\end{table*}

With the population value, search shows no gain over greedy on the players' later results (Table~\ref{tab:ladder}; $-$1.3, at 29\% coverage). The population judges, which see only the heroes, prefer it (56.7\%): search finds hero combinations the population model likes.

Search with a player-aware value and population proposals loses to player-aware greedy by 3.3 points. Its proposals still come from population behavior, and the prior was trained against a population value, so it is sharp: on 5,000 real pick states the player-aware value changes only 4.4\% of the population search's picks, while over a flat prior the same value changes 56\%.

At a fixed player-aware value, player-aware proposals change the result. They gain 4.5 points (2.1 to 7.0) on the paired record. The selected design adds ten-player bans, an opponent model and an assignment term (+3.1 against player-aware greedy), and the final drafter, with the support floor and calibration of Section~\ref{sec:support}, keeps +2.2 (0.4 to 3.9). A value network with player features predicts better but drafts no better (supplement).

\textit{Lookahead, budget or proposals?} The final drafter searches a tree, draws candidates from a learned player-aware prior with the ten-player terms, and spends 400 evaluations; greedy spends 4 rollouts on each of 30 fixed candidates. We first held the search to the root: each simulation picks a root candidate by the same rule, completes the draft with the same rollouts and applies the same value, so the tree below the root, and the opponent model in it, are removed. This simpler drafter is not detectably behind the full search ($+$0.3, $-$1.3 to 1.8) and beats player-aware greedy by 3.3 points (1.3 to 5.3), and by 2.4 (0.6 to 4.2) with greedy's budget of 120 evaluations. With no search at all, taking the top hero of the same root prior at every pick and ban, it is not detectably behind the full search ($-$0.4, $-$2.0 to 1.2) and still beats greedy by 2.2 (0.3 to 4.0). With a flat prior over greedy's 30 candidates and greedy's bans it loses 3.6 points (1.7 to 5.5) and is no better than greedy (+1.3, $-$0.4 to 3.0). Splitting that swap, greedy's 30 candidates ranked by the drafter's own prior cost nothing (+0.1, $-$1.3 to 1.5) and still beat greedy by 2.3 (0.6 to 4.1). Greedy's bans cost 0.9 ($-$0.6 to 2.3), and with them the lead over greedy is no longer detected (+0.9, $-$0.8 to 2.6). That prior is distilled from a 1,600-simulation search, so the finding is about decision time: search pays through the prior it trains, and searching again at play time adds nothing we can detect.

The rows change more than one part at a time and differ in coverage, so we rest on the direct contrasts: proposals (+4.5 and 3.6), lookahead (+0.3) and budget ($-$0.3) (Section~\ref{sec:limits}).

\textit{What population search gets wrong.} The population judges score population search at 63.4\% against the players' own drafts, but on the players' later results it is $-$1.6 ($-$3.9 to 0.7). It puts players on heroes they are worse at by 12.7 points of team skill against their own choices (population greedy: 7.4), giving away the value of players picking heroes they know.

\subsection{The final drafter}

\begin{table}[t]
\centering
\caption{The final drafter against each comparator on the 300 tournament lobbies, both seatings. Paired record as in Table~\ref{tab:ladder}. J: player-aware judge (\%); its 95\% CIs (lobby bootstrap, widened for search seeds) are within $\pm$1.2 points and listed in the supplement.}
\label{tab:tourn}
\setlength{\tabcolsep}{3pt}
\begin{tabular}{lccc}
\toprule
Final drafter vs & Paired record & Cov. & J \\
\midrule
Population search & \textbf{+5.4} (2.9, 7.9) & 37\% & 58.5 \\
Players' own drafts & \textbf{+2.8} (1.0, 4.6) & 57\% & 62.4 \\
Player-aware greedy & +2.2 (0.4, 3.9) & 53\% & 56.9 \\
Population greedy & +4.6 (2.2, 7.0) & 43\% & 60.4 \\
Imitation drafter & +5.6 (3.2, 8.0) & 47\% & 65.5 \\
DraftRec, pick head & +1.9 (0.2, 3.6) & 64\% & 57.3 \\
DraftRec, pick-and-win & +2.0 (0.1, 3.9) & 53\% & 54.7 \\
Selected design & +0.4 ($-$1.2, 2.1) & 58\% & 49.4 \\
\bottomrule
\end{tabular}
\end{table}

The final drafter beats every drafter that does not know the players, and every baseline, on the players' later results (Table~\ref{tab:tourn}). Covered players do 5.4 points better on its heroes than on population search's, and 2.8 points better than on the heroes they really played. Against the selected design, from which it differs only by the floor and the calibration, the paired record detects no difference ($+$0.4, $-$1.2 to 2.1). J uses the uncalibrated personal model, which favors the selected design slightly, so the paired record is the fair comparison between those two.

\textit{Selection protocol.} Designs were selected on 300 tune lobbies separate from the tournament, each by a rule fixed before it was applied. The selected design maximizes mean J against population search and search with player-aware value and proposals. The final drafter was admitted by non-inferiority to the selected design on the same tune lobbies ($-$0.39 points, $-$0.83 to 0.06; lower bound above $-$1.0 after correction for two candidates). The unconstrained selected design is reported beside it throughout.

\subsection{Robustness}
\label{sec:robust}

\textit{Further draws of lobbies.} The selected design was replayed on 300 new lobbies drawn under the same rule, disjoint from the tournament and tune lobbies. The judged results repeat (59.1\% and 62.5\%), and so does the paired record: +3.9 (1.5, 6.3) against population search and +2.4 (0.5, 4.2) against the players' own drafts. Pooled over all 771 lobbies that meet the rule in three disjoint draws, the record is +3.5 points (1.8, 5.1) against population search and +2.2 (0.8, 3.6) against the players' own drafts. On a looser draw of 300 lobbies that requires only 10 or more earlier games per player, the final drafter's paired record against population search is +1.0 ($-$1.6, 3.7), against +0.7 for the selected design. 

\textit{A judge with no shared fitting games.} A judge whose player model is refit on spring games only shares no fitting games with the drafters'. It is weaker, since the median lobby player has 2 spring games, but under it the selected design still scores 55.0\% (53.8, 56.1) against population search and 62.4\% against the players' own drafts.

\textit{A backtest on real games.} On 20,000 summer games, teams whose real pick matched the recommendation of a search with the player-aware value over the imitation prior won 0.82 points more (SE 0.36), controlling for the draft-only prediction and both teams' ratings. With the player model's predicted skill gap between the teams added as a control, the coefficient turns negative, $-$0.88 (SE 0.35): the gain runs through the skill gap the player model predicts, and at a fixed predicted gap, teams that matched the recommendation did slightly worse. This backtest uses an earlier search, not the final drafter. With the population prior in place of the imitation prior, it is $-$1.09 (SE 0.71) under the first controls. So the backtest checks the direction of the player model's estimates; it does not show that following the search wins more games.

\textit{Compute.} On one RTX 5090 a full draft takes 41 ms of search plus about 0.1 s for the opponent model.

\section{Staying on Support}
\label{sec:support}

\Pw{} showed that search over a hero-level value function leaves the data in composition space: most of its teams contain a three-hero combination humans have rarely played together. A player-aware drafter can also leave the data on a second axis, by putting players on heroes they have rarely played. Here the value function's player term rests on almost nothing. We measure both axes and constrain the first.

\subsection{Two constraints}

\textit{Composition floor.} The final drafter never makes a pick that creates a three-hero subset seen fewer than 25 times among the training teams. The 5th percentile of held-out human teams is 23, which we round to 25. The floor applies inside the search (at the root, in the tree, and in our rollout picks). If no hero in the team's pool passes it, the drafter takes the hero whose rarest triple is most common, so the floor gives way before the team leaves its players' pools.

\textit{Calibrated player term.} The player term is over-trusted (Section~\ref{sec:wrong}): on held-out games, only 0.78 of the term's claimed difference between teams shows up in results. A single scale of 0.864, set by a rule fixed in advance from the held-out slope on teams whose ten players are all known, multiplies the player weights in the drafter's value.

\subsection{Composition support}

The selected design, unconstrained, puts 38\% of its teams below the floor. The final drafter puts 1.3\% there, against 6.3\% of the humans' own teams in the same lobbies. It stays close to the floor rather than deep inside the data: the median count of its rarest triple is 50, against 201 for humans. On the players' later results the paired record detects no difference between the final drafter and the unconstrained design ($+$0.4, $-$1.2 to 2.1). This contrast moves the floor and the calibration together, so it does not isolate the floor. \Pw{} found that a play-time floor costs real value unless the prior is trained under it. Here it costs little, because the players' pools already keep the search nearer the data: 38\% of the selected design's teams are below the floor before it is applied, against 63\% for hero-level search, so the floor binds less often. A variant whose prior is trained under the floor, as in \pwof{}, does no better in truth (+0.80 against +0.95 for the final drafter in the hero-only worlds of Section~\ref{sec:truth}).

\subsection{Player support}

\begin{table}[t]
\centering
\caption{Player support: earlier ranked games of each placed player on the hero they are given, 300 tournament lobbies against a common panel of opponents (imitation drafter, the players' own drafts, DraftRec). Drafters' players are placed by earlier play counts (the paired record's rule); ``same rule'' places the humans' heroes the same way. 95\% CIs by lobby bootstrap.}
\label{tab:familiar}
\setlength{\tabcolsep}{3pt}
\begin{tabular}{lccc}
\toprule
Drafter & Never played & Under 5 games & Median \\
\midrule
Humans (as played) & 3.2\% & 13.1\% (11.9, 14.3) & 41 \\
Humans (same rule) & 2.3\% & 9.4\% (8.4, 10.4) & 47 \\
Final drafter & 3.8\% & 9.9\% (9.0, 10.8) & 59 \\
Selected design & 3.7\% & 9.2\% (8.5, 10.1) & 63 \\
\begin{tabular}[t]{@{}l@{}}Player-aware value\\ \quad and proposals\end{tabular} & 4.5\% & 10.7\% (9.9, 11.5) & 59 \\
Player-aware greedy & 5.1\% & 12.6\% (11.7, 13.6) & 44 \\
\begin{tabular}[t]{@{}l@{}}Player-aware value,\\ \quad population proposals\end{tabular} & 17.9\% & 32.2\% (30.9, 33.4) & 16 \\
Population search & 18.5\% & 33.2\% (31.9, 34.5) & 15 \\
\bottomrule
\end{tabular}
\end{table}

The drafters whose proposals know the players stay on the player axis without being told to (Table~\ref{tab:familiar}). Fewer than 5 earlier games on a hero is the point below which the player model treats it as new. Under the same placement rule, the final drafter puts 9.9\% of its players there and the humans' own teams 9.4\%, though it places more on heroes never played (3.8\% against 2.3\%); its median player has 59 earlier games on their hero against 47. Population search goes far off this axis: 18.5\% of its players are on a hero they have never played, a third are under 5 games, and the median is 15. So does search with a player-aware value and population proposals, which explains its loss in Table~\ref{tab:ladder}: the value function knows those placements are bad, but the proposals keep offering them.

Across the ladder, the searches that leave the player axis do not beat greedy and those that beat it stay on the axis; this does not isolate the effect of familiarity.

\subsection{What the floor costs in truth}
\label{sec:truth}

The paired record is noisy, so we also price the floor in simulated worlds where the true value of every draft is known. We extend the ground-truth worlds of \pwof{} with a player layer. Each world gives every player a true effect on every hero: a familiarity effect by earlier games on the hero, a general skill beyond the rating that can drift by half-year, a role effect, and a player-hero cell, fit jointly with the hero-level model on 2.03 million ranked games before the judge window and drawn from its posterior. The layer is deliberately built differently from the drafters' player model, with no borrowing across heroes, so the drafters' estimates are imperfect measurements of a known truth. Players take a team's heroes in the arrangement that is best in that world (construction, checks and the tune-lobby results in the supplement). We use three families of worlds. In the \emph{hero-only} worlds of \pwof{}, players have no effect on outcomes. In \emph{f-cal}, the player term is fit on game outcomes alone, using the same raw records and scoreboards the drafters see but a different model (a regularized regression on shrunk record and scoreboard summaries, 322,332 games), with none of the drafters' estimates. In \emph{f-cov}, it also takes the drafters' own skill estimates as a fitted covariate, so it is partly circular. On teams whose rarest triple was seen 25 to 99 times, outcomes follow 0.72 of the drafters' personal term in f-cal and 0.77 in f-cov, against 0.74 in real games; f-cal's scale was set to reach that slope, and f-cov contains the drafters' estimates, so this match is partly by construction. The independent check is that f-cal predicts held-out real games better than the drafters' own model (log loss 0.6570 against 0.6652).

Against the unconstrained selected design, the floor alone changes true value by $-$0.04 ($-$0.33 to 0.27) in the hero-only worlds, $-$0.25 ($-$0.56 to 0.15) in f-cal and $-$0.17 ($-$0.46 to 0.23) in f-cov. The final drafter, floor and calibration together, gains +0.95 (0.67 to 1.22) in the hero-only worlds and shows no detected difference in either player family ($-$0.22, $-$0.50 to 0.12, in f-cal; $-$0.21, $-$0.49 to 0.12, in f-cov). In worlds where players matter as much as they do in real games, staying on support costs nothing we can detect, and gains nothing either. An outcome-only layer fit without the scoreboard follows the personal term at only 0.20 and favors the final drafter by +0.64; we report it in the supplement and do not rely on it.

\section{Where the Value Function Is Still Wrong}
\label{sec:wrong}

\textit{The player term is over-trusted.} On held-out real teams, 0.78 of the player term's claimed difference shows up in results. After the single scale set in advance, the share is 0.89 on all teams and 0.96 when all ten players are known. What remains is an experience tilt: the share is 0.75 for the least experienced quarter of players and 1.04 for the most experienced. Records of newer players still promise more than they deliver. A forced switch also costs more than the model predicts: when a player's best hero is banned or taken (16\% of slots), they do 1.99 points worse than the draft predicts, against 1.45 predicted, an under-prediction of 0.54 points (SE 0.13).

\textit{The players do not explain the hero part's interaction error.} \Pw{} found that the hero-level value function claims about 10\% too much team-interaction value. A missing player factor is not the cause: with the player term in the model, the share that shows up rises only from 0.84 to 0.85, and the two terms are uncorrelated across teams (0.01).

\textit{The gap to a perfect value function.} Giving the final drafter the true value of every draft raises its true win probability against population search, over its learned value, by 8.1 points in the hero-only worlds, 4.9 (4.4 to 5.6) in f-cal and 3.6 in f-cov. Swapping in the truth one part at a time puts most of that gap in the player term, and most of the player term's share in noise no estimator could recover from the records: a player's true skill on a lightly played hero is not in their data.

\section{Drafting for Part of a Team}
\label{sec:party}

Most players queue alone or in a party of two or three: 46\% of teams are five solo players and 25\% hold one duo. A party controls only its own picks, so we built a party drafter and measured what it is worth.

\textit{Rules and setup.} In ranked drafts the team's highest-ranked player makes the bans, and on each team pick turn whoever locks first takes the turn, so a player's pick turn is a choice and their ban control depends on their rank in the lobby (here their Heroes Profile rating). The party uses the final drafter. In the simulation each stranger and opponent picks by sampling DraftRec for that player at temperature 1, the strangers locking in an order the drafter does not know; in the drafter's search they are chance nodes with DraftRec's mixture over the players still to pick, and a ban the party does not hold is sampled from a model of the holder. We compare the party's picks with typical picks (the party picking like strangers), with the population search drafter on the same seats and with the heroes the party really played, sharing the party, pick order and random draws across arms. The tournament's judges score the final drafts on all 1,175 tournament-window lobbies that meet the tournament's rule outside the tune and tournament lobbies.

\subsection{Gain by party size}

\begin{table}[t]
\centering
\caption{Party drafter: judged gain (points of the party team's win probability, J) over typical picks, share of the full team's gain, J over the population search drafter on the same seats, and the paired record (points per player on the party's team with later games on their hero). Gains average over the lock turns as drawn. 1,175 lobbies, 95\% CIs.}
\label{tab:party}
\setlength{\tabcolsep}{2.6pt}
\begin{tabular}{lcccc}
\toprule
Party & Gain (J) & Share & vs pop.\ search & Record \\
\midrule
Solo & +2.2 (1.9, 2.6) & 20\% & +2.6 & +0.4 ($-$0.3, 1.1) \\
Duo & +4.2 (3.7, 4.6) & 36\% & +4.9 & +0.1 ($-$0.8, 0.9) \\
Trio & +7.2 (6.7, 7.7) & 62\% & +7.6 & +0.6 ($-$0.3, 1.4) \\
Four & +9.4 (8.9, 9.9) & 82\% & +10.4 & $-$0.3 ($-$1.3, 0.6) \\
Five & +11.5 (10.9, 12.0) & 100\% & +12.8 & +0.6 ($-$0.3, 1.5) \\
\bottomrule
\end{tabular}
\end{table}

A solo player gains 2.2 points of judged win probability over picking like a typical player (Table~\ref{tab:party}). That is a fifth of what a full team of five gains (+11.5). A duo gets 36\%, a trio 62\% and a four-stack 82\%, so the gain grows roughly in proportion to the number of seats the party controls. Against the population search drafter on the same seats the gains are larger at every size (+2.6 to +12.8), for the same reason as in the tournament. Against the heroes the party members really played, the drafter adds 1.9 points for a solo player and 8.9 for a full team.

\subsection{When to lock, and who bans}

\textit{Lock last.} In simulation, a solo player's gain rises with each later pick turn up to the team's fourth, and the fourth and last turns are within noise of each other, with or without the bans (supplement). Weighted by how often solo players hold the bans, the gain rises from 1.6 points on the team's first pick turn to 2.8 on the last. Locking first instead of last costs 1.1 points of gain over typical picks (0.7 to 1.5). Seeing more picks outweighs the risk of losing the player's best hero (still free in 74 to 78\% of drafts at the last turn, 90 to 92\% at the first). An adaptive drafter that decides at each turn whether to wait, from its own value estimates, does no better than always locking last ($-$0.3 points, $-$0.6 to 0.0).

\textit{Holding the bans.} A solo player holds the bans 19.0\% of the time in real lobbies, a duo 43.4\%, a trio 66.1\% and a four-stack 84.7\%. Holding them adds 2.3 points for a solo player and 2.1 for a duo.

\textit{Other judges and the paired record.} Against the population search drafter the paired record is positive at every party size, from +1.0 (0.5 to 1.6) for a solo player to +4.2 (3.0 to 5.5) for a full team. Against typical picks it is small and its intervals include zero at every size (+0.6, $-$0.3 to 1.5, for a full team), so the judged gains over typical picks are not confirmed player by player.

\section{Keeping It Current}
\label{sec:upkeep}

Both halves of the value function age. Measured as the correlation of true values between two 90-day windows a year apart, corrected for luck~\cite{spearman1904proof}, a hero's win rate in a tier keeps 0.85 and a player's skill on a hero they play often keeps 0.74. A week-old player record loses a fifth of the players' 5.91-point gain in accuracy. In a replayed 2.8-year deployment, retraining the draft model every three to six builds keeps all but 0.02 to 0.06 of the 1.00 point an unmaintained model loses. So the draft side needs retraining a few times a year and the player records daily. A test on game builds released after September 27, 2026 is registered (\osfurl). The supplement gives the full measurements.

\section{Limitations}
\label{sec:limits}

\textit{The paired record is not a win rate.} The primary measure reads players' later results on heroes they chose themselves, for the 29 to 64\% of players covered under both drafters, mostly established players. It tests whether the player edges a drafter builds on persist. In the observational backtest, which uses an earlier search, matching the recommendation goes with winning only through the skill gap the player model predicts, and slightly with losing at a fixed gap. Outcomes of games played with the drafter's recommendations measure the drafter directly; a prospective test of this kind, with recommendations shown to players and their games recorded, is under way.

\textit{The ladder is a pattern.} We do not test directly whether the search-against-greedy gap differs between rows, and the stages of the ladder change more than one part at a time. The direct contrasts of proposals, lookahead and budget are the tested evidence. The split of the proposal swap compares each version with the same root-held drafter, not with each other; no arm pairs a flat prior with the drafter's own bans, and only the flat-prior swap shows a loss that excludes zero. The prior-only interval still allows the full search up to 2.0 points more.

\textit{The value function is still wrong in known ways.} The player term remains over-trusted for newer players (a slope of 0.75 for the least experienced quarter after calibration), and the worlds put most of the remaining gap in skill the records cannot reveal.

\textit{Support.} The final drafter sits close to the composition floor rather than deep inside the data (median rarest triple 50 against 201 for humans). The ground-truth worlds encode a model class. Both player-world families reproduce the real calibration of the player term, but f-cov does so by including the drafters' own estimates, and f-cal shares the drafters' raw inputs.

\textit{Judges and simulated teammates.} Judged results come from models. In the party study, strangers and opponents do not adapt to the party, and the paired record confirms the gains over population search but not over typical picks.

\textit{Scope.} These are ranked games of one game. Professional play has scheduled opponents and prepared strategies, and newer players (under 50 earlier games) are covered by the judge more than by the paired record.

\section{Data, Code and Ethics}
\label{sec:ethics}

All data are public match records from Heroes Profile's public interface, collected within its terms of service, which do not allow the records to be redistributed; we share none of them. Code and models are released at \repourl. Our analyses identify players only by numeric account identifiers, and no player name appears in the paper. The study observed public records only and involved no contact with players. Any deployment that scouts opponents uses only what the game client already shows in a lobby, matched to public records.

\section{Conclusion}
\label{sec:conclusion}

Adding the players to a standard draft model raises accuracy on an untouched later patch from 57.2\% to 62.7\%. Search over a hero-only value shows no detected gain over greedy drafting, and adding the players to the value alone is not enough: the search must also propose heroes the players can play. When both the value and the proposals know the players, the drafter wins on the players' later results, by 2.2 points per covered player over player-aware greedy and 2.8 over the players' own drafts. Picking straight from its learned prior, with no search at play time, it is not detectably behind; with a flat prior over greedy's candidates it loses its edge. For drafting with players, how the learned prior ranks the heroes matters more than how deep or how long the search looks at play time. The final drafter stays on support on both axes, composition and the player on each hero, with no detected cost.

\section*{Acknowledgment}
We thank Heroes Profile (heroesprofile.com), whose public interface supplied our replay corpus and player histories. Heroes of the Storm is a trademark of Blizzard Entertainment.

Generative AI (Claude Code, Anthropic; Codex, OpenAI) has been used in the preparation of portions of this document, both in manuscript editing as well as in code writing, design, and data analysis. The authors directed the research, chose the experiments and framing, checked the results and text, and take full responsibility for the content.

\bibliographystyle{IEEEtran}
\bibliography{references}

@misc{paper1,
  author = {Segan, Max and Pysher, Ernest and Frankel, James and Lipson, Hod},
  title = {Searching Beyond the Data: Value-Function Exploitation and the Cost of Staying on Support in {MOBA} Drafting},
  note = {Under review, IEEE Transactions on Games, manuscript ToG-2026-0267},
  year = {2026}
}

@inproceedings{draftrec,
  author = {Lee, Hojoon and Hwang, Dongyoon and Kim, Hyunseung and Lee, Byungkun and Choo, Jaegul},
  title = {{DraftRec}: Personalized Draft Recommendation for Winning in Multi-Player Online Battle Arena Games},
  booktitle = {Proc. ACM Web Conference (WWW)},
  year = {2022}
}

@inproceedings{chen2018art,
  author = {Chen, Zhengxing and Nguyen, Truong-Huy D. and Xu, Yuyu and Amato, Christopher and Cooper, Seth and Sun, Yizhou and El-Nasr, Magy Seif},
  title = {The Art of Drafting: A Team-Oriented Hero Recommendation System for Multiplayer Online Battle Arena Games},
  booktitle = {Proc. ACM Conference on Recommender Systems (RecSys)},
  year = {2018}
}

@inproceedings{summerville2016draft,
  author = {Summerville, Adam and Cook, Michael and Steenhuisen, Ben},
  title = {Draft-Analysis of the Ancients: Predicting Draft Picks in {DotA} 2 Using Machine Learning},
  booktitle = {Proc. AAAI Conf. Artificial Intelligence and Interactive Digital Entertainment (AIIDE) Workshops},
  year = {2016}
}

@article{gourdeau2021discriminative,
  author = {Gourdeau, Daniel and Archambault, Louis},
  title = {Discriminative Neural Network for Hero Selection in Professional {Heroes of the Storm} and {DOTA} 2},
  journal = {IEEE Transactions on Games},
  volume = {13},
  number = {4},
  pages = {380-387},
  year = {2021}
}

@article{chen2021juewudraft,
  author = {Chen, Sheng and Zhu, Menghui and Ye, Deheng and Zhang, Weinan and Fu, Qiang and Yang, Wei},
  title = {Which Heroes to Pick? {L}earning to Draft in {MOBA} Games with Neural Networks and Tree Search},
  journal = {IEEE Transactions on Games},
  volume = {13},
  number = {4},
  pages = {410-421},
  year = {2021}
}

@misc{berner2019dota,
  author = {Berner, Christopher and others},
  title = {Dota 2 with Large Scale Deep Reinforcement Learning},
  note = {arXiv:1912.06680},
  year = {2019}
}

@article{hodge2021win,
  author = {Hodge, Victoria J. and Devlin, Sam and Sephton, Nick and Block, Florian and Cowling, Peter I. and Drachen, Anders},
  title = {Win Prediction in Multiplayer Esports: Live Professional Match Prediction},
  journal = {IEEE Transactions on Games},
  volume = {13},
  number = {4},
  pages = {368-379},
  year = {2021}
}

@inproceedings{kim2017team,
  author = {Kim, Young Ji and Engel, David and Woolley, Anita Williams and Lin, Jeffrey Yu-Ting and McArthur, Naomi and Malone, Thomas W.},
  title = {What Makes a Strong Team? {U}sing Collective Intelligence to Predict Team Performance in {League of Legends}},
  booktitle = {Proc. ACM Conf. Computer-Supported Cooperative Work and Social Computing (CSCW)},
  year = {2017}
}

@article{sapienza2018individual,
  author = {Sapienza, Anna and Zeng, Yilei and Bessi, Alessandro and Lerman, Kristina and Ferrara, Emilio},
  title = {Individual Performance in Team-Based Online Games},
  journal = {Royal Society Open Science},
  volume = {5},
  number = {6},
  pages = {180329},
  year = {2018}
}

@book{elo1978rating,
  author = {Elo, Arpad E.},
  title = {The Rating of Chessplayers, Past and Present},
  publisher = {Arco},
  year = {1978}
}

@article{glickman1999parameter,
  author = {Glickman, Mark E.},
  title = {Parameter Estimation in Large Dynamic Paired Comparison Experiments},
  journal = {Journal of the Royal Statistical Society: Series C (Applied Statistics)},
  volume = {48},
  number = {3},
  pages = {377-394},
  year = {1999}
}

@inproceedings{herbrich2006trueskill,
  author = {Herbrich, Ralf and Minka, Tom and Graepel, Thore},
  title = {{TrueSkill}: A {B}ayesian Skill Rating System},
  booktitle = {Advances in Neural Information Processing Systems},
  year = {2006}
}

@article{efron1975data,
  author = {Efron, Bradley and Morris, Carl},
  title = {Data Analysis Using {S}tein's Estimator and Its Generalizations},
  journal = {Journal of the American Statistical Association},
  volume = {70},
  number = {350},
  pages = {311-319},
  year = {1975}
}

@article{brown2008inseason,
  author = {Brown, Lawrence D.},
  title = {In-Season Prediction of Batting Averages: A Field Test of Empirical {B}ayes and {B}ayes Methodologies},
  journal = {Annals of Applied Statistics},
  volume = {2},
  number = {1},
  pages = {113-152},
  year = {2008}
}

@article{spearman1904proof,
  author = {Spearman, Charles},
  title = {The Proof and Measurement of Association between Two Things},
  journal = {American Journal of Psychology},
  volume = {15},
  number = {1},
  pages = {72-101},
  year = {1904}
}

@article{kalman1960new,
  author = {Kalman, Rudolf E.},
  title = {A New Approach to Linear Filtering and Prediction Problems},
  journal = {Journal of Basic Engineering},
  volume = {82},
  number = {1},
  pages = {35-45},
  year = {1960}
}

@inproceedings{koren2009temporal,
  author = {Koren, Yehuda},
  title = {Collaborative Filtering with Temporal Dynamics},
  booktitle = {Proc. ACM SIGKDD Int. Conf. Knowledge Discovery and Data Mining},
  year = {2009}
}

@article{koren2009matrix,
  author = {Koren, Yehuda and Bell, Robert and Volinsky, Chris},
  title = {Matrix Factorization Techniques for Recommender Systems},
  journal = {Computer},
  volume = {42},
  number = {8},
  pages = {30-37},
  year = {2009}
}

@article{bradley1952rank,
  author = {Bradley, Ralph Allan and Terry, Milton E.},
  title = {Rank Analysis of Incomplete Block Designs: {I}. {T}he Method of Paired Comparisons},
  journal = {Biometrika},
  volume = {39},
  number = {3/4},
  pages = {324-345},
  year = {1952}
}

@article{silver2018general,
  author = {Silver, David and others},
  title = {A General Reinforcement Learning Algorithm that Masters Chess, Shogi, and {Go} through Self-Play},
  journal = {Science},
  volume = {362},
  number = {6419},
  pages = {1140-1144},
  year = {2018}
}

@article{rosin2011multi,
  author = {Rosin, Christopher D.},
  title = {Multi-Armed Bandits with Episode Context},
  journal = {Annals of Mathematics and Artificial Intelligence},
  volume = {61},
  number = {3},
  pages = {203-230},
  year = {2011}
}

@article{browne2012survey,
  author = {Browne, Cameron B. and others},
  title = {A Survey of {M}onte {C}arlo Tree Search Methods},
  journal = {IEEE Transactions on Computational Intelligence and AI in Games},
  volume = {4},
  number = {1},
  pages = {1-43},
  year = {2012}
}

@article{gama2014survey,
  author = {Gama, Jo{\~a}o and {\v{Z}}liobait{\.e}, Indr{\.e} and Bifet, Albert and Pechenizkiy, Mykola and Bouchachia, Abdelhamid},
  title = {A Survey on Concept Drift Adaptation},
  journal = {ACM Computing Surveys},
  volume = {46},
  number = {4},
  year = {2014}
}

@book{quinonero2009dataset,
  editor = {Qui{\~n}onero-Candela, Joaquin and Sugiyama, Masashi and Schwaighofer, Anton and Lawrence, Neil D.},
  title = {Dataset Shift in Machine Learning},
  publisher = {MIT Press},
  year = {2009}
}

@inproceedings{sculley2015debt,
  author = {Sculley, D. and Holt, Gary and Golovin, Daniel and Davydov, Eugene and Phillips, Todd and Ebner, Dietmar and Chaudhary, Vinay and Young, Michael and Crespo, Jean-Fran{\c{c}}ois and Dennison, Dan},
  title = {Hidden Technical Debt in Machine Learning Systems},
  booktitle = {Advances in Neural Information Processing Systems},
  year = {2015}
}

@inproceedings{do2021player,
  author = {Do, Tiffany D. and Wang, Seong Ioi and Yu, Dylan S. and McMillian, Matthew G. and McMahan, Ryan P.},
  title = {Using Machine Learning to Predict Game Outcomes Based on Player-Champion Experience in {League of Legends}},
  booktitle = {Proc. 16th Int. Conf. Foundations of Digital Games (FDG)},
  year = {2021}
}

@inproceedings{kim2016proficiency,
  author = {Kim, Jooyeon and Keegan, Brian C. and Park, Sungjoon and Oh, Alice},
  title = {The Proficiency-Congruency Dilemma: Virtual Team Design and Performance in Multiplayer Online Games},
  booktitle = {Proc. CHI Conf. Human Factors in Computing Systems},
  pages = {4351--4365},
  year = {2016}
}

@techreport{minka2018trueskill2,
  author = {Minka, Tom and Cleven, Ryan and Zaykov, Yordan},
  title = {{TrueSkill} 2: An Improved {Bayesian} Skill Rating System},
  institution = {Microsoft Research},
  number = {MSR-TR-2018-8},
  year = {2018}
}

@article{lu2019drift,
  author = {Lu, Jie and Liu, Anjin and Dong, Fan and Gu, Feng and Gama, Jo{\~a}o and Zhang, Guangquan},
  title = {Learning under Concept Drift: A Review},
  journal = {IEEE Trans. Knowl. Data Eng.},
  volume = {31},
  number = {12},
  pages = {2346--2363},
  year = {2019}
}

@inproceedings{hanke2017recommender,
  author = {Hanke, Lucas and Chaimowicz, Luiz},
  title = {A Recommender System for Hero Line-Ups in {MOBA} Games},
  booktitle = {Proc. AAAI Conf. Artificial Intelligence and Interactive Digital Entertainment (AIIDE)},
  pages = {43--49},
  year = {2017}
}

@inproceedings{semenov2016performance,
  author = {Semenov, Aleksandr and Romov, Peter and Korolev, Sergey and Yashkov, Daniil and Neklyudov, Kirill},
  title = {Performance of Machine Learning Algorithms in Predicting Game Outcome from Drafts in {Dota 2}},
  booktitle = {Analysis of Images, Social Networks and Texts (AIST), CCIS vol. 661},
  pages = {26--37},
  year = {2016}
}

\end{document}


\title{Supplementary Material for\\ Drafting for the Players: Player-Aware Values and Proposals for MOBA Draft Search}
\author{Max~Segan, Ernest~Pysher, and Hod~Lipson%
\thanks{M. Segan is with the Department of Computer Science and H. Lipson with the Department of Mechanical Engineering, Columbia University (e-mail: m.segan@columbia.edu; hod.lipson@columbia.edu). E. Pysher is an independent researcher (e-mail: ernestpysheriv@gmail.com).}}

\maketitle

\section{Player model specification}
\label{sec:s-spec}

This section specifies the full player model behind the main text's headline numbers. Stages run in the order given; each stage after the factor model is fit on the calibration window (the games after the draft model's training cutoff, which falls during February 10, through March 28, 2026) with earlier stages frozen. Table~\ref{tab:s-spec} lists what fits each stage.

\textit{Notation.} Slot $i$ is player $p$ on hero $h$ in a game on day $t_i$; $y_i = 1$ if the player's team won. Probabilities are fractions (points $\times 100$ in the prose). For a game on day $t$, $n_{ph}$ and $n_p$ count the player's games on the hero and overall on earlier days, corpus and Heroes Profile together; $\tilde n_{ph}, \tilde n_p$ count corpus games only; $\Delta_{ph}$ is the days since the player last played the hero.

\subsection{The record}

For a corpus game, $\pi_i$ is the draft-only model's win probability for the player's team, $r_i = y_i - \pi_i$ and $v_i = \pi_i(1-\pi_i)$. For a game known only from Heroes Profile, $\pi_i$ is the hero's win rate in the game's calendar quarter, $(W_{hQ} + 200 b_h)/(N_{hQ} + 200)$ with $b_h = (W_h + 100)/(N_h + 200)$, from fitting-window corpus games and clipped to $[0.2, 0.8]$; later quarters reuse the last fitted one.

An experience offset is subtracted from every residual, $\tilde r_i = r_i - \mu(\tilde n_p, \tilde n_{ph})$, at that row's own counts. It is the mean residual of corpus games from April 1, 2024 to February 10, 2026 in $8 \times 8$ count bins (overall 0, 1, 5, 10, 20, 50, 100, 300+; on the hero 0, 1, 2, 3, 5, 10, 20, 50+), shrunk to zero with 200 pseudo-games. It runs from $-6.7$ points (300+ games, hero never played) to $+4.0$.

The record of $p$ on $h$ is kept in information form, $S_{ph} = \sum \tilde r_i/v_i$ and $P_{ph} = \sum 1/v_i$, so the record is $\bar y_{ph} = S_{ph}/P_{ph}$ with noise $1/P_{ph}$. Its level drifts as a random walk with variance $q_w$ per day. Before a day's rows enter a cell, and again at scoring time,
\begin{equation}
(S_{ph}, P_{ph}) \leftarrow (S_{ph}, P_{ph}) / (1 + q_w P_{ph} \delta), \label{eq:s-drift}
\end{equation}
with $\delta$ the days since the cell last changed: the Kalman filter of a local-level model, which keeps the record's mean and lowers its precision. With $q_w = 0.02^2/365$ (2 points a year), an $n$-game record keeps half its precision for $1/(q_w P) \approx v/(q_w n)$ days, about 6 years at 100 games and 2 at 300. Drifts of 0.5 to 8 points a year were compared on the calibration window; 1.5 to 3 score within 0.06 thousandths of log loss.

Scoreboard numbers $z_{im}$ (Section~\ref{sec:s-norm}) accumulate the same way, $X_{phm} = \sum z_{im}$ over the $C_{phm}$ games with a value, discounted by Eq.~\eqref{eq:s-drift} with precision $C_{phm}/\varepsilon_m^2$ and $q_m = 0.4^2/365$. At 100 games a profile keeps half its precision for 2 days (regeneration globes) to 5 months (experience), median 10 days. The six team shares (hero damage, damage taken, healing, siege, experience, takedowns) use precision $C_{phm}$.

\subsection{Overall level and learned hero traits}

Skill of $p$ on $h$, in residual units, is
\begin{equation}
\theta_{ph} = \mathbf u_h^\top \mathbf a_p + e_{ph}, \quad \mathbf a_p \sim N(\mathbf 0, I_3), \quad e_{ph} \sim N(0, s_h^2),
\end{equation}
with $\mathbf u_h = (\sigma_0, V_{h1}, V_{h2})$: an overall level, equal on every hero, and two learned hero traits. Scoreboard means $\bar x_{phm} = X_{phm}/C_{phm}$ load on the same $\mathbf a_p$: $\bar x_{phm} - c_{hm} \sim N(\mathbf g_m^\top \mathbf a_p, \rho_m^2 + \varepsilon_m^2/C_{phm})$, with $\mathbf g_m$ shared by heroes and zero where a number is dropped. The posterior of $\mathbf a_p$ has precision $\Lambda_p$ and shift $\boldsymbol\eta_p$,
\begin{align}
\Lambda_p &= I + \textstyle\sum_h w_{ph} \mathbf u_h \mathbf u_h^\top + \sum_{h,m} \omega_{phm} \mathbf g_m \mathbf g_m^\top, \\
\boldsymbol\eta_p &= \textstyle\sum_h w_{ph} \bar y_{ph} \mathbf u_h + \sum_{h,m} \omega_{phm} (\bar x_{phm} - c_{hm})\, \mathbf g_m,
\end{align}
with $w_{ph} = P_{ph}/(1 + s_h^2 P_{ph})$ and $\omega_{phm} = C_{phm}/(\rho_m^2 C_{phm} + \varepsilon_m^2)$.

The parameters are fit on a fixed random half of the players (105,319) by maximum predictive likelihood: given each player's evidence before March 1, 2025 (no drift), the exact Gaussian density of their corpus-game records from March 1, 2025 to February 10, 2026. Gaussian penalties have standard deviation 0.05 on $V$, 0.3 on $\mathbf g_m$ and $c_{hm}$, and 0.5 on $\log s_h^2$ around its mean; L-BFGS runs $3 \times 500$ iterations. The fit gives $\sigma_0 = 5.65$ points, trait columns of root-mean-square 3.67 and 1.39 points, and $s_h$ of 2.14 to 8.20 points (median 3.18).

\subsection{The trust curve}

The prior for the scored hero is the posterior with that hero's own record removed ($S_{ph} = P_{ph} = 0$):
\begin{equation}
m_0 = \mathbf u_h^\top \Lambda_p^{-1}\boldsymbol\eta_p, \qquad v_0 = \mathbf u_h^\top \Lambda_p^{-1} \mathbf u_h + s_h^2 .
\end{equation}
It combines the player's overall level, their records on heroes with similar traits and their scoreboard on every hero. A Gaussian model gives the record weight $w = P_{ph}/(P_{ph} + 1/v_0)$. The full player model uses
\begin{equation}
\hat\theta = \beta_{\mathrm p}(\mathbf z)\, m_0 + \beta_{\mathrm o}(\mathbf z)\, w (\bar y_{ph} - m_0) + \beta_{\mu}(\mathbf z)\, \mu, \label{eq:s-trust}
\end{equation}
each $\beta(\mathbf z) = \mathbf z^\top \boldsymbol\beta$ linear in $\mathbf z = (1, \ell, \ell^2, \sigma, \gamma, \ell\sigma, \sigma^2, \gamma^2, \ell\gamma)$, where $\ell = \ln(1+n_{ph})/\ln 301$, $\sigma = n_{ph}/\max(n_p, 1)$ and $\gamma = \ln(1+\Delta_{ph})/\ln 1001$ ($0$ if never played). The 27 coefficients are fit by least squares of $r_i - \mu$ on calibration-window slots, weights $1/v_i$. On those slots $\beta_{\mathrm o}$ averages 2.32 at 1 to 9 games on the hero, 1.45 at 10 to 29, 1.09 at 30 to 99, 0.96 at 100 to 299 and 1.04 at 300+, against Gaussian weights $w$ of 0.02, 0.07, 0.17, 0.33 and 0.53. Write $s^{(1)} = \mu + \hat\theta$.

\subsection{Short-term layer}

With $u_j = r_j - s^{(1)}_j$ for the player's earlier corpus games on any hero,
\begin{equation}
\phi_p = \frac{\sum_j 2^{-(t - t_j)/45}\, u_j}{\sum_j 2^{-(t - t_j)/45} + 160}, \qquad \nu_p = \frac{\sum_{j \in E} u_j}{|E| + 5},
\end{equation}
where $E$ holds games played with fewer than 5 earlier games on their hero, and $s^{(2)} = s^{(1)} - 0.169\,\phi_p + 0.097\,\nu_p \mathbf 1[n_{ph} < 5]$. Half-life (45, 180, 365, 730 days) and pseudo-counts (10, 40, 160; 1, 2, 5, 20) were chosen by calibration-window log loss of a game-level logistic regression on the draft-only logit and the team differences of $s^{(1)}$, $\phi$ and $\nu$; the two weights are its coefficients divided by the one on $s^{(1)}$ (4.37). A recent run above expectation is partly given back.

\subsection{Direct scoreboard channel}

From the discounted state the channel reads, for each of the 15 numbers and 6 team shares, $X_{phm}/(C_{phm} + 5)$ on the hero and $\sum_{h'} X_{ph'm}/(\sum_{h'} C_{ph'm} + 5)$ across heroes: 42 averages, standardized on the calibration window to $\mathbf f_i$. Then $s^{(3)} = s^{(2)} + \mathbf f_i^\top(\mathbf b + \mathbf d_{k(h)})$, with $k(h)$ one of six roles (bruiser, healer, melee assassin, ranged assassin, support, tank), minimizing on calibration-window slots
\begin{equation}
\textstyle\sum_i (r_i - s^{(2)}_i - \mathbf f_i^\top(\mathbf b + \mathbf d_{k}))^2/v_i + \lambda\lVert \mathbf b\rVert^2 + \lambda_d \sum_k \lVert \mathbf d_k\rVert^2 .
\end{equation}
$\lambda \in \{10^2, \dots, 10^7\}$ and then $\lambda_d \in \{10^3, \dots, 10^8, \infty\}$ were chosen by a two-fold split of calibration-window players: $\lambda = 10^5$, $\lambda_d = 10^6$.

\subsection{Calibration and the game}

The slot estimate is $s = s^{(3)}\, \mathbf b(\mathbf z)^\top \mathbf c$, where $\mathbf b(\mathbf z)$ is $\mathbf z$ plus hinges $(x - x_0)_+$ in $\ell$ at 100 and 300 games, in $\sigma$ at 0.6 and 0.8, and in $\gamma$ at 180 and 365 days. The 15 coefficients are fit by least squares of $r_i$ on $s^{(3)}_i \mathbf b_i$, weights $1/v_i$; the multiplier's 5th to 95th percentiles are 0.85 to 1.03 (Table~\ref{tab:s-cal} gives the calibration).

The rating term uses each player's latest Heroes Profile rating before the game. Slot covariates $\mathbf x_i$ are the team's mean rating minus the opponents', the player's minus their team's, the player's hero rating minus their rating (each per 100) and a known-hero-rating indicator, all zero unless all ten ratings are known. A logistic regression with offset $\operatorname{logit}\pi_i$ on corpus games from April 1, 2024 to February 10, 2026 gives $\boldsymbol\kappa = (0.046, 0.002, 0.021, 0.090)$ and $a_i = \operatorname{logit}^{-1}(\operatorname{logit}\pi_i + \boldsymbol\kappa^\top\mathbf x_i) - \pi_i$. For teams $A$ and $B$,
\begin{equation}
\operatorname{logit}\Pr(A \text{ wins}) = c_0 + c_1 \operatorname{logit}\pi_A + c_2 D_s + c_3 D_a,
\end{equation}
where $D_s = \sum_A s - \sum_B s$ and $D_a = \sum_A a - \sum_B a$. Maximum likelihood on the calibration window gives $(c_0, \dots, c_3) = (0.0002, 1.012, 4.337, 0.421)$.

\subsection{Scoring a game}

For a game on day $t$, player-history rows precede $t$ (a Heroes Profile row is dated by its game's end) and ratings use the last pregame value:
\begin{enumerate}
\item For each player, add earlier rows to the cells, discounting each touched cell by Eq.~\eqref{eq:s-drift} to the row's day; discount all cells to $t$. Read $n_{ph}, n_p, \Delta_{ph}, \tilde n_{ph}, \tilde n_p$ and $\mu$.
\item Compute $m_0, v_0$ without the scored hero's record, then $w$, $\mathbf z$ and $s^{(1)}$ (Eq.~\eqref{eq:s-trust}).
\item Add the layer from earlier $u_j$, each from that game's own as-of $s^{(1)}_j$; add the direct channel; apply the calibration multiplier.
\item Combine the ten slots' $s$ and $a$ by team with the draft-only prediction.
\end{enumerate}
A game's own result and scoreboard enter only the states of later days.

\begin{table}[h]
\centering
\caption{What fits each stage. Fitting window: corpus games up to the draft model's training cutoff during February 10, 2026. Calibration: games after that cutoff, through March 28, 2026 (70,974 games); no game is in both.}
\label{tab:s-spec}
\setlength{\tabcolsep}{3pt}
\begin{tabular}{lll}
\toprule
Stage & Fit on & Settings \\
\midrule
History baseline & fitting window & quarters, 200 pseudo-games \\
Experience offset & Apr 2024--Feb 2026 & 64 bins, 200 pseudo-games \\
Scoreboard & Apr 2024--Feb 2026 & Section~\ref{sec:s-norm} \\
Rating term & Apr 2024--Feb 2026 & 4 covariates \\
Factor model & half of players & rank 2 + overall level \\
Drift & calibration & 2 points, 0.4 SD a year \\
Trust curve & calibration & 27 coefficients \\
Short-term layer & calibration & 45 days; 160, 5 \\
Direct channel & calibration & $\lambda = 10^5$, $\lambda_d = 10^6$ \\
Calibration & calibration & 15 coefficients \\
Game combiner & calibration & 4 coefficients \\
\bottomrule
\end{tabular}
\end{table}

\section{Provenance and leakage controls}
\label{sec:s-prov}

Table~\ref{tab:provenance} lists, for each component of the prediction model, the games it is fit on, how its inputs are dated, and how it updates. Player records and scoreboard profiles exclude the predicted day; ratings use the last value before the game starts. The draft-only weights are trained on the fitting-window games whose residuals the player model learns from, which shrinks those residuals somewhat; the history-only baseline is a hero's win rate over its calendar quarter (a strictly as-of version changes accuracy by at most 0.08 points). Match histories were pulled after each window and are dated by game end, so a record can include a game played before the predicted day but uploaded after it.

\begin{table}[h]
\centering
\caption{Provenance of the prediction model's components. Fit: the games its parameters are estimated on. Inputs: how the information it reads is dated.}
\label{tab:provenance}
\setlength{\tabcolsep}{2.5pt}
\small
\begin{tabular}{p{0.30\columnwidth}p{0.27\columnwidth}p{0.36\columnwidth}}
\toprule
Component & Fit & Inputs; update \\
\midrule
Draft-only weights & through Feb 10 & frozen \\
Draft-only statistics & -- & earlier builds; each build \\
History-only baseline & through Feb 10 & hero's quarter; static \\
Scoreboard baselines & through Feb 10 & last fitted quarter; static \\
Hero traits & through Feb 10 & frozen \\
Combining weights, trust curve, calibration & Feb 10 to Mar 28 & frozen \\
Player records, profiles & -- & games before day $t$; daily \\
Ratings & -- & last value before the game \\
\bottomrule
\end{tabular}
\end{table}

Spring and summer games supported model selection for the trust curve, team shares and role weights, so their gains are development estimates. The post-freeze set is the confirmatory evaluation: it touched no step of model selection, and we scored it once with every weight frozen and player records updating daily. Of 95,973 patch 2.57 games through October 26, 7,341 with the new hero and 3,974 with a missing player row drop out, leaving 84,658 (88\%). In the drafter tournament, the judge's and the drafters' player models are fit on disjoint halves of the players but share game outcomes; the main text's robustness checks refit the judge without them.

\section{Scoreboard normalization}
\label{sec:s-norm}

Each scoreboard number goes through six steps:
\begin{enumerate}
\item It becomes a per-minute rate and is compressed with an inverse hyperbolic sine, which tames the few huge games and keeps zeros.
\item It is standardized against games on the same hero and map, in the same one of five game-length bands, with the same result. Uther's healing is measured against other Uther games. Comparing only games with the same result removes the part of the number that comes from winning or losing, so the scoreboard adds information the win record does not already carry. Where a hero has few games on a map, the baseline blends in the hero's all-map baseline as if it were 100 extra games. Baselines use fitting-window games only.
\item It is capped at 5 standard deviations, so a disconnect cannot dominate.
\item It gets a per-hero, per-quarter adjustment, fit on fitting-window games, so a balance patch (a game update that makes heroes stronger or weaker) that raised everyone's damage on a hero in that window does not look like the player changing. Quarters after the fitting window reuse the last fitted adjustment, so later patches are not removed.
\item Numbers that carry no information for a hero, such as healing on a hero with no heals, are dropped for that hero.
\item The player's share of the team's damage, damage taken, siege, experience, takedowns and healing enters alongside.
\end{enumerate}

\textit{Alternatives.} Each alternative replaced one step of the normalization, was refit, and was scored against the base normalization on the spring and summer test sets, paired by game (Table~\ref{tab:s-norm}). The chosen normalization, shrinking thin hero-map baselines toward the hero's all-map baseline, leaves accuracy unchanged and predicts heroes a player has rarely played much better: on held-out cold starts the share of real variation explained rises from 0.24 to 0.35.

\begin{table}[h]
\centering
\caption{Each alternative against the base normalization: accuracy difference (points, SE) on the spring and summer test sets, and log-loss gain (thousandths), spring / summer. Positive favors the alternative. Measured with the base player model.}
\label{tab:s-norm}
\setlength{\tabcolsep}{3pt}
\begin{tabular}{lccc}
\toprule
Change & Spring & Summer & Log loss \\
\midrule
Hero-map shrinkage (used) & +0.06 (0.08) & $-$0.02 (0.04) & +0.07 / +0.05 \\
Tier-split baselines & $-$0.04 (0.09) & $-$0.06 (0.05) & $-$0.45 / $-$0.48 \\
Smooth length curves & +0.08 (0.09) & $-$0.10 (0.05) & $-$0.24 / $-$0.37 \\
Share of team totals only & $-$0.05 (0.10) & $-$0.15 (0.05) & $-$0.44 / $-$0.76 \\
Difference from opponent & $-$0.09 (0.10) & $-$0.21 (0.05) & $-$1.42 / $-$1.34 \\
Rank-based scores & +0.06 (0.07) & 0.00 (0.03) & +0.32 / +0.27 \\
Cap at 2.5 & +0.08 (0.06) & +0.01 (0.03) & +0.18 / +0.17 \\
Weight by decisiveness & +0.10 (0.07) & 0.00 (0.04) & +0.13 / $-$0.10 \\
\midrule
No scoreboard at all & $-$1.26 (0.15) & $-$1.40 (0.07) & $-$8.16 / $-$8.46 \\
\bottomrule
\end{tabular}
\end{table}

Replacing a player's numbers with their share of team totals, or with the difference from the opposing player, removes the player's absolute level, which is the part that carries to the next game. Added alongside the absolute numbers in the full model, team shares gain 0.57 thousandths of log loss on the summer test set (week-block 95\% CI 0.40 to 0.73).

\section{Scoreboard reliability, all numbers}

\begin{table}[h]
\centering
\caption{Stable share of one game's number for one player on one hero, how many times the win or loss it is, and the games needed before half a record is real.}
\label{tab:s-rel}
\setlength{\tabcolsep}{4pt}
\begin{tabular}{lccc}
\toprule
Number & Stable share & Times win/loss & Games \\
\midrule
Win or loss, beyond the draft & 0.013 & 1 & 74 \\
Protection given (17 heroes) & 0.40 & 30 & 1.5 \\
Self-healing & 0.39 & 29 & 1.6 \\
Damage taken & 0.28 & 21 & 2.6 \\
Siege damage & 0.28 & 21 & 2.6 \\
Hero damage & 0.23 & 17 & 3.4 \\
Healing (35 heroes) & 0.23 & 17 & 3.4 \\
Experience contribution & 0.21 & 16 & 3.8 \\
Regeneration globes & 0.20 & 15 & 4.1 \\
Deaths & 0.17 & 13 & 4.8 \\
Mercenary captures & 0.16 & 12 & 5.2 \\
Structure damage & 0.14 & 10 & 6.4 \\
Deaths while outnumbered & 0.09 & 7 & 9.9 \\
Kills & 0.07 & 5 & 14 \\
Escapes & 0.06 & 4 & 17 \\
Assists & 0.04 & 3 & 23 \\
\bottomrule
\end{tabular}
\end{table}

Games to half real is $(1 - r)/r$ by the Spearman-Brown rule, with $r$ the stable share.

\section{Checks of the carry-forward numbers}

\begin{table}[h]
\centering
\caption{Share of a player's past record on a hero that showed up in their next game on it, by past games on the hero (spring and summer test sets). Noise only: the share expected if true skill never changed. Persistence: measured divided by noise only.}
\label{tab:s-carry}
\setlength{\tabcolsep}{4pt}
\begin{tabular}{lccc}
\toprule
Past games on the hero & Measured & Noise only & Persistence \\
\midrule
30 to 59 & 0.24 & 0.30 & 0.80 \\
60 to 119 & 0.30 & 0.38 & 0.79 \\
120 to 249 & 0.43 & 0.50 & 0.85 \\
250 to 499 & 0.50 & 0.62 & 0.81 \\
500 to 799 & 0.58 & 0.70 & 0.83 \\
800 or more & 0.81 & 0.79 & 1.02 \\
\bottomrule
\end{tabular}
\end{table}

\textit{Simulation.} With true skill that never changes (spread 3, 4 or 5 points), the estimator returns the noise-only prediction within 0.03 in every games bin. With true skill that drifts at a correlation of 0.74 a year, it falls to about 0.65 of the noise-only prediction in the heaviest bins. The real data stay at 0.8 or above, between the two.

\textit{Recency.} Records of the last 25, 50 or 100 games, and records with half-lives of 180 or 365 days, predict the next game worse than the full record. Recent games carry slightly more real signal but are much noisier.

\textit{Specialists.} Implied true persistence by the hero's share of the player's games: under 15\%, 0.91 to 0.96 from 30 to 799 games; over 50\%, 0.56 at 120 to 249 games and 0.66 at 250 to 499. Adding at-game ratings changes these by under 0.1.

\textit{Variance of one game.} The draft explains 3.4\% of one game's variance. Each player's stable overall effect beyond the lobby's rating is 0.27\% (2.7\% for all ten), all of it from players with under 300 earlier games. Each player's stable hero-specific effect is 0.45\% (4.5\% for ten). Under this additive approximation, about 89\% remains unexplained. Out of sample, the players add 5.3 points of explained variance on all 358,490 test games, and 4.2 on the 5,120 games in which all ten players have 300 or more earlier games.

\section{Carry-forward, worked example}

Records about 12 points above expectation over 70 to 90 games realized 3.8 points in the next game, 32\% of the record. The same records over 200 to 260 games realized 6.6 points, 57\%. Losing records shrink the same way: $-$11.7 over 70 to 90 games realized $-$3.5, and $-$11.5 over 200 to 260 realized $-$6.9. 

\section{Keeping it current, in full}
\label{sec:s-upkeep}

Both halves of the value function age: hero strength through balance patches, and player skill through practice and time away. This section gives the full measurements behind the main text's upkeep section: how fast each half changes, what falling behind costs, and how often each half needs refreshing.

\subsection{How much the game and the players change}
\label{sec:s-clocksec}

Balance patches move hero strength, and players get better or worse on their heroes. We measure both sides with one estimator, in the same unit (points of win probability), on the same games.

\textit{Measuring fairly.} We estimate every cell from two separate 90-day windows, the second starting 3 months to 2 years after the first ends. A game-side cell is a hero in one of three tier groups, valued as its win rate minus 50. A player-side cell is a player on a hero, valued as their results minus the draft-only prediction, less their own average over all heroes. Each window's value is the true value plus luck of known size, so inverse-variance pooling gives the correlation of the true values (Spearman's correction for attenuation), with bootstrap intervals; window pairs start every 30 days and are pooled.

\begin{figure}[h]
\centering
\includegraphics[width=\columnwidth]{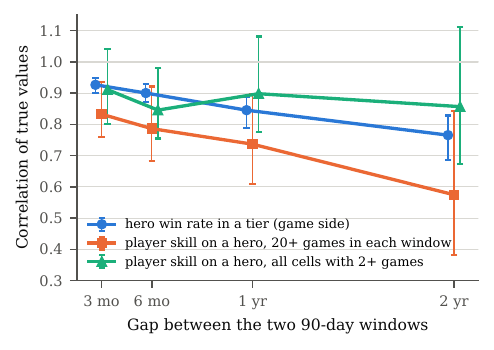}
\caption{The two clocks: correlation of true values between two separate 90-day windows, by the gap between them, with 95\% bootstrap CIs. Values in Table~\ref{tab:s-clocks}.}
\label{fig:s-clocks}
\end{figure}

\textit{The game.} A hero's win rate in a tier keeps 0.93 after 3 months, 0.85 after a year and 0.77 after 2 years (Fig.~\ref{fig:s-clocks}, Table~\ref{tab:s-clocks}). Patches drive this clock. Heroes the patch notes left alone keep 0.95 at 3 months and 0.91 at a year, while heroes changed twice or more keep 0.75 to 0.80.

\textit{The players.} Across player-hero cells with at least two games in both windows, pooled correlations are 0.85 to 0.91 at every gap, with wide intervals. On heroes played 20 or more times in each window, the heroes players rely on, skill keeps 0.83 at 3 months, 0.74 at a year and 0.57 at 2 years.

\textit{Reading.} Point estimates suggest comparable change in hero strength and in skill on frequently played heroes, and a player's edge on a hero (true spread 3.3 points) is larger than a hero's tier edge (2.4 points). The player model's drift setting, 1.5 to 3 points a year, agrees: a player's edge on a hero changes on a scale of years, about as fast as the game.

\subsection{What a year without updates costs a prediction}

We use two sets. To compare the two halves, we left each alone for up to a year and scored it on the same 60,500 tournament-window games, whose players all have 50 or more earlier games (Table~\ref{tab:s-stale}). To measure the player side alone, we lagged the player records of the full model on the 286,016 summer test games (Table~\ref{tab:lag}). The draft model trained through June 2, 2025 is about 12 to 13 months out of date for these games; the current one is trained through February 10, 2026.

\begin{table}[h]
\centering
\caption{Accuracy (\%) with each half up to a year out of date, 60,500 tournament-window games, full player model. Rows: age of the draft model. Columns: no players, current player records, or records as old as the draft model.}
\label{tab:s-stale}
\setlength{\tabcolsep}{4pt}
\begin{tabular}{lccc}
\toprule
Draft model & None & Current records & Records same age \\
\midrule
Current & 56.84 & 62.74 & 62.74 \\
4 months old & 56.34 & 62.56 & 58.87 \\
1 year old & 55.62 & 62.18 & 57.84 \\
\bottomrule
\end{tabular}
\end{table}

\textit{The draft side.} On its own, a year-old draft model loses 1.22 of its 6.84 points of edge (18\%). With current player records it loses 0.56, because the player terms make up part of it. All 90 heroes were known a year earlier, and every hero-tier cell already had at least 3,779 games (median 37,919), so none of the loss comes from missing heroes or thin cells. It comes from change in heroes and matchups the model already knew.

\textit{Both sides.} With everything a year old, accuracy falls from 62.74\% to 57.84\%: the edge drops from 12.74 to 7.84 points, a loss of 4.90 (38\%). Measured by log loss, the full model keeps 23\% of its lift over a coin flip after a year, against 72\% for the draft-only model.

\textit{The player side, lag by lag.} On the summer test set the players add 5.91 points of accuracy (56.94\% to 62.85\%). Player records one week old already lose 1.16 of them, a fifth. A month old, they lose 2.38 (40\%), and a year old, 4.36 (74\%). In log loss the losses are a quarter, a half and nine tenths of the players' gain (Table~\ref{tab:lag}). A record a month old does worse, in log loss, than the win-only model run on current records.

\begin{table}[h]
\centering
\caption{Full player model with lagged player records, 286,016 summer test games: accuracy, and the log-loss cost against current records (thousandths, share of the players' gain of 37.38).}
\label{tab:lag}
\setlength{\tabcolsep}{4pt}
\begin{tabular}{lcc}
\toprule
Player records & Accuracy (\%) & Log-loss cost \\
\midrule
Current (day before) & 62.85 & \\
1 week old & 61.69 & 9.45 (25\%) \\
1 month old & 60.47 & 20.02 (54\%) \\
4 months old & 59.33 & 28.20 (75\%) \\
1 year old & 58.49 & 33.49 (90\%) \\
\midrule
Draft only & 56.94 & \\
\bottomrule
\end{tabular}
\end{table}

\textit{Why the player side loses more.} Giving only one group of player slots its year-old record, players with no games a year earlier (23.4\% of slots) cost about half of the total, heroes a known player had never played (15.9\%) a fifth, and known player-hero pairs (60.7\%) a quarter. About three quarters of the loss is people and heroes a year-old record has never seen, a gap the draft model does not have.

\subsection{Upkeep for each half}

\textit{The draft side: retrain a few times a year.} A build is one game client version; builds came about a month apart. We replayed a 2.8-year deployment of the draft-only model forward across 36 builds, from about 640,000 games of history in July 2023. At each build a maintenance policy either retrains the model on all games so far, recomputes its statistics tables from earlier builds (a refresh), or leaves both alone, and the deployed model is scored on the next build's games. Each policy's regret is its games-weighted accuracy loss, averaged over the whole deployment, against retraining after every build (Section~\ref{sec:s-drift}); every policy is replayed with three training seeds, and we report the mean. A draft model never maintained loses 1.00 points of accuracy (0.97 to 1.02 across seeds), 15\% of its 6.87-point edge. Refreshing the statistics at every build with the weights fixed recovers 0.36 of that in every seed. Retraining every three builds, with the statistics refreshed in between, recovers nearly all of it (0.02 left), and every six builds leaves 0.06, so a cadence of three to six builds, quarterly to twice a year, keeps the draft side current.

\textit{The player side: update records every day.} The cost of a stale player model arrives fast: against records current to the day before, a week's lag costs a fifth of the players' accuracy gain and a month's lag 40\% (Table~\ref{tab:lag}). The post-freeze test scored the frozen weights on one new patch with no refit, and they kept their lead with a calibration slope of 0.956; the slope should be checked on each patch and the weights refit when it moves.

\subsection{Upkeep for the search drafter}

Against a year-old twin built from out-of-date data, population search scores 53.3\% (51.8, 54.8) under J and 52.2\% under the population judges (51.8\% under J at four months). Fine-tuning the stale twin for 2 GPU-hours leaves the current drafter's lead at 52.2\%, and refreshing its statistics alone at 53.2\%; both sit inside the interval of the unmaintained lead, so this test cannot rank the maintenance options. With only the player records a year old, the player-aware search with player proposals scores 53.7\% (53.2, 54.2) against its stale twin under J. The current drafter's paired record is 1.2 points higher than the year-old drafter's (0.9 to 1.5, widened for search seeds). A test of this lead on game builds released after September 27, 2026 is registered (\osfurl).

\section{The trust curve and calibration}
\label{sec:s-cal}

\begin{table}[h]
\centering
\caption{Calibration slope of realized results on predictions, read from a smooth curve over games on the hero at each listed point (spring and summer test sets; post-freeze set in the last column). Near 1,000 games few slots remain: the post-freeze value there is 0.73 (0.51, 0.98).}
\label{tab:s-cal}
\setlength{\tabcolsep}{4pt}
\begin{tabular}{lccc}
\toprule
Games on hero & Base model & Full model & Post-freeze \\
\midrule
0 & 1.31 & 1.05 & 0.84 \\
3 & 1.16 & 0.99 & 0.87 \\
10 & 1.11 & 0.99 & 0.91 \\
30 & 0.99 & 0.95 & 0.94 \\
100 & 0.88 & 1.00 & 0.92 \\
300 & 0.76 & 0.89 & 1.01 \\
1000 & 0.80 & 0.84 & 0.73 \\
\bottomrule
\end{tabular}
\end{table}

The trust curve multiplies the weight a Gaussian model gives a record by 2.32 at 1 to 9 games, 1.45 at 10 to 29, 1.09 at 30 to 99, 0.96 at 100 to 299 and 1.04 at 300 or more (Section~\ref{sec:s-spec}). A record of a player's first games on a hero deserves about twice the weight the static prior gives it, which fits new-hero results carrying more player-specific signal than a fixed prior assumes. At 1,000 or more games and after 600 to 1,000 days without play, records are still trusted slightly too much.

\section{Tournament details}

\textit{Lobby rule.} Real games with a full 16-step draft whose ten picks join to known players, all ten with 50 or more earlier games and 10 or more judge-window games; 300 lobbies drawn with a fixed seed. The tune set is a separate 300 lobbies.

\textit{Further draws.} Three more disjoint draws of 300 lobbies, with the same pipeline, settings and intervals, exclude the tournament, tune and drafter-fitting lobbies: a replicate under the same rule; a looser draw requiring 10 or more earlier games (171 of its lobbies also meet the main rule, and 129 have a player under 50 games); and a newer-player draw requiring at least one player under 50 earlier games in every lobby (median player 565 earlier games; 13.7\% of players under 50). Under J the four draws agree (heterogeneity $p \geq 0.38$ for every matchup). The selected design's paired record against population search, real-draft replay, the imitation drafter and DraftRec's pick head is +4.8, +3.1, +3.6, +2.1 on the tournament lobbies; +3.9, +2.4, +5.1, +3.0 on the replicate; $-$0.3, +0.1, $-$0.1, +0.9 on the looser draw's 171 lobbies that meet the main rule; and +1.6, +0.3, +4.6, +1.8 on the newer-player draw. Inverse-variance pools: over the 771 lobbies meeting the main rule, +3.5 (1.8, 5.1), +2.2 (0.8, 3.6), +3.4 (1.8, 4.9), +2.1 (1.0, 3.2) with two-way intervals (random effects against population search: +3.1, 0.5 to 5.7); the remaining pools use lobby-bootstrap intervals: over the 429 newer-player lobbies, +1.7 ($-$0.6, 3.9), +1.0 ($-$0.8, 2.7), +3.8 (1.7, 5.9), +1.5 (0.0, 2.9); over all 1,200 lobbies, +2.8 (1.6, 4.1), +1.8 (0.8, 2.8), +3.6 (2.4, 4.7), +1.9 (1.0, 2.7).

\textit{Paired-record uncertainty.} Players recur across lobbies (in the tournament draw 2,146 distinct players fill 3,000 slots; 579 hold 1,433) and across draws, and their later games are reused. We therefore cluster the paired record by lobby and by player: the two-way variance is $V_\text{lobby} + V_\text{player} - V_\text{slot}$, with a slot the intersection of a lobby and a player (Cameron, Gelbach and Miller, J.\ Bus.\ Econ.\ Stat., 2011), each from 2,000 bootstrap draws and floored at the larger one-way variance; the pooled figures bootstrap the three draws jointly with shared player weights. The one-way SEs are within 8\% of each other, because a recurring player rarely meets the same pair of heroes again (4--9\% of covered slots repeat a player and hero pair), so the two-way intervals are within 8\% of the lobby-bootstrap ones; in this data a product-weight (pigeonhole) bootstrap widens them by about $\sqrt{3}$. Without subtracting the draft-only expectation, the tournament-draw differences are +4.5, +4.0, +5.5, +3.4, +3.0 and +0.6 points (main tournament table order), with the same signs; pooled over the three draws, +2.7, +3.4, +5.1 and +3.4 against population search, real drafts, the imitation drafter and DraftRec. Covered players, against uncovered ones: about 1,300 against 800 earlier games, twice the judge-window games, 41 against 5 earlier games on the comparator's hero (against population search), rating about 35 points higher, the same share of solo players (53\%).

\textit{Bradley-Terry strengths.} Under J (logit scale, imitation drafter at 0) the selected design leads at 0.726 (0.705, 0.746), ahead of player-prior search (0.663), DraftRec (0.428 to 0.543), population search (0.339) and real-draft replay (0.196).

\textit{Combiner.} The drafters' personal value $V$ is fit on 5,397 games that the draft-only value network never trained on (February 11 to May 22, 2026).

\textit{Judge.} Fit on 144,414 judge-window games: personal coefficient 3.05 (SE 0.04), log-loss lift over the population consensus 23.2 thousandths (SE 0.55). \textit{Population judges.} All but one population judge are fit on 260,539 games: 136,517 replays of the drafters' period uploaded after their snapshot and 124,022 games of the build that spans May 11 to July 20, 2026, which include 299 of the 300 tournament lobbies. The Quick Match judge is fit on 116,589 Quick Match games from July 2025 to June 22, 2026 and contains none of them. With it as the only population part of J (personal term unchanged), the selected design scores 58.0\% against population search, 64.0\% against real drafts, 65.3\% against the imitation drafter, 59.0\% against DraftRec's pick head, 57.3\% against its pick-and-win rule and 51.2\% against the player-prior search, each with its lobby-bootstrap interval above 50. Used alone in place of the consensus, the Quick Match judge moves every population-judge cell of the tournament table by at most 2.4 points and none crosses 50 (43.4\% against population search, 59.9\% against real drafts, 58.0\% against the imitation drafter, 58.2\% and 53.4\% against DraftRec's two rules, 48.4\% against the player-prior search); on the second draw of lobbies it gives 43.8\%, 59.5\%, 57.5\% and 58.1\%. In the party study it puts the gain over typical picks at 1.80 points for a solo player and 8.64 for a full team, against 1.56 and 7.55 with the consensus; under J with this population part the gains are 2.66 and 13.39, with team shares of 20\%, 39\%, 65\% and 82\%.

Recalibrated on held-out games, J\textsubscript{cal} weights the consensus by 1.18 and the personal term by 1.00, so the personal term is calibrated and the consensus is underconfident.

\begin{table}[h]
\centering
\caption{Three judges: the original, one with every fitted coefficient refit on spring games (partial), and one refit entirely on spring games (full). Selected design's J win \% against each opponent.}
\label{tab:s-judges}
\setlength{\tabcolsep}{4pt}
\begin{tabular}{lccc}
\toprule
& Original & Partial & Full \\
\midrule
Personal coefficient & 3.05 & 3.00 & 2.76 \\
Log-loss lift ($\times$1000) & 23.2 & 23.4 & 17.5 \\
Scale (judge per drafter unit) & 0.55 & 0.56 & 0.44 \\
\midrule
vs population search & 59.0 & 59.4 & 55.0 \\
vs real-draft replay & 63.0 & 63.0 & 62.4 \\
vs imitation drafter & 66.0 & 65.9 & 65.0 \\
vs DraftRec, pick head & 58.1 & 57.9 & 58.7 \\
vs DraftRec, pick-and-win & 55.1 & 55.0 & 54.8 \\
vs player-prior search & 51.6 & 51.7 & 50.8 \\
\bottomrule
\end{tabular}
\end{table}

Under the full (disjoint-fit) judge, player-prior search scores 54.6\% against population search, against 58.2\% under the original judge; its personal-sum scale is also lower. Contrasts carried by population win rates barely move. Some near-even matchups below the top flip, for example DraftRec against population search (52.8\% to 48.1\%). The earlier search drafter against real drafts rises to 56.7\%.

\textit{DraftRec.} Rebuilt from the paper and its released code: a player transformer over each player's last 50 games with scoreboard statistics and a draft transformer, with pick and win heads. On the summer test set its pick head is right on 36.1\% of next picks (top five: 68.9\%), and its post-draft win prediction has an AUC of 0.652. DraftRec scores games with a full 16-step draft order on a map and tier it knows: 55,612 of the 72,474 spring games and 203,666 of the 286,016 summer games. For the prediction comparison, DraftRec is trained with the same code and settings on 749,711 drafts from April 1, 2024 to February 9, 2026, its win head reaches 60.51\% accuracy on those spring games and 60.16\% on summer, gaining 15.09 (12.77, 16.55) and 15.58 thousandths of log loss over the draft-only model, with AUC 0.648 and 0.643. On the same games the full player model reaches 63.15\% and 62.77\%, gaining 36.00 and 37.25, and the win-records-only model gains 18.9 to 19.7.

\section{Design study, full}

\begin{table}[h]
\centering
\caption{Four ways to put the players into the search, on the 300 tournament lobbies: J win \% against a search with a population prior (population search for a, b, d and the selected design; the behavioral-prior population search for a$'$ and c), and the training the player part needs.}
\label{tab:s-options}
\setlength{\tabcolsep}{3pt}
\begin{tabular}{p{3.6cm}cp{3.2cm}}
\toprule
Option & J & Training \\
\midrule
(a) Leaf only, MCTS-trained prior & 50.9 & none \\
(a') Leaf only, behavioral prior & 56.3 & none \\
(b) Prior with player features & 58.2 & 0.4 GPU-hours of targets, 1 CPU-minute fit \\
(c) Value network with player features, behavioral prior & 56.8 & CPU minutes \\
(d) Self-play over real rosters & 56.1 & 1.1 GPU-hours \\
\midrule
Selected design, vs the earlier drafter & 59.0 & as (b), plus opponent model \\
\bottomrule
\end{tabular}
\end{table}

Options (a) and (d) face population search; (a$'$) and (c) face the weaker behavioral-prior search. Self-play over real rosters (d) scores 49.6\% under the population judges. Added to the player-prior search, the refit value network scores 50.0\% (49.6, 50.4) against the same search without it.

\begin{table}[h]
\centering
\caption{Each change alone against the player-prior search on the tournament lobbies, J and population-judge win \% (95\% CIs for J), and the tune-lobby score used for selection. The table lists one entry per change and the three combinations; the three combinations score 55.28 to 55.40. Selection used selection-stage player models and the design was then held fixed; the scores shown use the evaluation player models. The comfort opponent models each opponent's picks from the heroes that opponent plays most, in place of DraftRec (mean J against population search and the player-prior search on 300 separate lobbies).}
\label{tab:s-design}
\setlength{\tabcolsep}{2.5pt}
\resizebox{\columnwidth}{!}{%
\begin{tabular}{lccc}
\toprule
Change & J & Population & Tune \\
\midrule
\textbf{Selected design} (DraftRec opponent) & 51.6 (51.1, 52.1) & 48.8 & 55.29 \\
Combined, comfort opponent & 52.0 (51.5, 52.4) & 48.6 & 55.40 \\
Combined, 1,600 simulations & 51.6 (51.0, 52.2) & 49.0 & 55.28 \\
Bans at opponents' best heroes & 51.7 (51.2, 52.2) & 49.2 & 54.94 \\
Opponent comfort model & 50.3 (49.8, 50.8) & 49.9 & 54.23 \\
Assignment bias & 49.5 (49.1, 49.9) & 49.5 & 53.87 \\
Refit value network & 50.0 (49.6, 50.4) & 49.7 & 53.86 \\
Prior weight by pick position & 49.1 (48.7, 49.6) & 52.1 & 53.96 \\
1,600 simulations & 49.8 (49.4, 50.2) & 50.5 & 53.62 \\
Exploration constant 4 & 49.5 (49.0, 50.0) & 49.9 & 53.76 \\
Exploration constant 1 & 48.3 (47.9, 48.7) & 49.4 & 51.86 \\
Personal bias halved & 48.9 (48.4, 49.3) & 51.6 & 53.47 \\
Personal bias doubled & 49.6 (49.0, 50.2) & 47.1 & 53.37 \\
More weight on the population MCTS-trained prior & 48.7 (48.1, 49.3) & 52.9 & 53.73 \\
No weight on the population MCTS-trained prior & 47.7 (46.8, 48.6) & 45.9 & 51.35 \\
Optimism, half an sd & 49.8 (49.4, 50.2) & 50.1 & 53.82 \\
Optimism, one sd & 48.9 (48.1, 49.6) & 49.9 & 52.46 \\
Rich prior, search targets & 46.8 (46.3, 47.4) & 47.3 & 50.88 \\
Rich prior, outcome-weighted & 47.6 (47.1, 48.1) & 45.5 & 51.39 \\
Rich prior, one-step targets & 47.9 (47.4, 48.4) & 48.8 & 51.44 \\
\midrule
No change (control) & 50.0 (49.3, 50.8) & 50.1 & 53.95 \\
\bottomrule
\end{tabular}}
\end{table}

\section{Implementation and verification of the search}

The tree search runs on the GPU, one thread block per search and 256 threads sharing each network evaluation. The selected design's DraftRec opponent runs once on the host at each decision and enters the GPU search as a fixed bias. Our nodes choose by the PUCT rule (predictor plus upper confidence bound), $Q(a) + c\,P(a)\sqrt{N}/(1 + n(a))$ with $c = 2$. At opponent nodes the reply is drawn from the opponent model, with progressive widening that allows at most $\lceil\sqrt{N}\rceil$ distinct replies at a node with $N$ visits. The tree is closed-loop: every node is one exact draft state, so the search looks through the opponent's replies to our later picks. A finished draft is valued once, in both team orders, and the value is backed up from the searching team's side. Simulations run one at a time inside a block, so a search is fully determined by its inputs and seed.

The CPU replica ports the GPU's random number generator and reproduces it on 64 of 64 subsequences and 8,192 first draws. Across 512 randomized and adversarial positions in the four player-aware search variants, 504 searches matched visit for visit through 400 simulations, and each of the other 8 first diverged within a simulation that contained a near tie. Under other exploration and widening settings, 252 of 256 searches matched and four diverged at a near tie; overall, 756 of 768 matched. On 256 full eight-turn episodes of the earlier search, 248 were identical. Best assignments matched brute force on 400 trials. The opponent model's draws pass chi-square tests (8,192 draws). A search on small drafts with exact leaf values converges to the optimum as simulations grow, with mean regret 0.181, 0.125, 0.044, 0.024, 0.0025 and 0.0022.

\section{Search budget and compute}

Against population search on the tune lobbies, the selected design gains 2.2 points from 100 to 400 simulations (56.4\% to 58.6\%) and is flat after that. Population search does not gain at any budget (49.8\% to 50.2\%), because its prior already puts almost all visits on one or two heroes.

On one RTX 5090, a full draft of eight decisions at 400 simulations takes 34.8 ms for population search, 35.9 ms with the leaf-only player term and 41.1 ms with the player prior. The selected design adds host-side inference for the DraftRec opponent, about 0.10 seconds per draft in all. Computing the ten players' estimates costs about 0.1 ms per player-day. The earlier work's self-play prior took 8.3 GPU-hours per seed; the player prior adds 0.4 GPU-hours of search targets and a minute of CPU fitting.

\section{The final drafter held to the root}

The root-held control is the final drafter with its tree capped at the root. It keeps the same GPU kernel, root prior (player-aware prior with the assignment and denial-ban terms), support floor and role mask at the root, PUCT constant, leaf value (draft-only value plus the scaled personal terms), 400 simulations per decision and choice rule (most visits). Each simulation chooses a root hero by PUCT and finds no further node, so it completes the draft with the search's rollout policy (the imitation drafter under the pool masks for both teams, with the floor on our side) and applies the leaf value. Every root candidate is therefore scored by rollouts, with as many value evaluations per decision as the full search. What is removed is the tree below the root: our later decisions are not searched, and the DraftRec opponent model at opponent nodes is replaced by the rollout policy. Player-aware greedy, by contrast, takes the top 15 heroes by the imitation model and the top 15 by personal value, gives each 4 rollouts (120 evaluations) and picks the best.

\begin{table}[h]
\centering
\caption{The root-held control on the 300 tournament lobbies, both seatings. Paired record as in the main text (points per covered player, 95\% CI); J and the population-judge consensus in \%.}
\label{tab:s-matched}
\setlength{\tabcolsep}{3pt}
\begin{tabular}{lcccc}
\toprule
Root-held vs & Paired record & Cov. & J & Pop.\ judges \\
\midrule
Final drafter & $-$0.3 ($-$1.8, 1.3) & 58\% & 49.8 & 50.1 \\
Selected design & +0.4 ($-$1.0, 1.8) & 59\% & 49.2 & 49.9 \\
Player-aware value and proposals & +1.1 ($-$0.4, 2.5) & 59\% & 50.8 & 48.7 \\
Player-aware greedy & +3.3 (1.3, 5.3) & 53\% & 56.4 & 53.4 \\
Population greedy & +2.0 ($-$0.1, 4.2) & 43\% & 60.5 & 52.1 \\
Population search & +2.6 (0.0, 5.2) & 37\% & 58.3 & 44.3 \\
Players' own drafts & +1.8 (0.0, 3.6) & 56\% & 62.1 & 58.9 \\
Imitation drafter & +4.1 (1.8, 6.3) & 46\% & 65.5 & 58.7 \\
\bottomrule
\end{tabular}
\end{table}

Its teams fall below the composition floor in 2.1\% of drafts (final drafter 1.3\%).

\textit{Splitting the gain over greedy.} Two further arms swap one part of the root-held control for greedy's. With greedy's budget, it runs 120 simulations per decision instead of 400 and keeps its own proposals. With a flat prior on greedy's candidates and bans, the root sees only greedy's 30 candidates (the top 15 by the imitation drafter and the top 15 by the team's best personal term among the free heroes of the team's pools) under a flat prior with no ten-player root terms, the bans are the imitation drafter's, and it keeps 400 simulations. A second round splits the candidate swap: greedy's bans alone; greedy's bans with the team-assignment term removed from the root prior; and greedy's 30 candidates ranked by the drafter's own prior and terms, with its own bans. A third round removes the search: at every pick and ban the drafter takes the top hero of the root prior the search would start from (the learned player-aware prior with the assignment and denial-ban terms, under the support floor and structural mask). Everything else is unchanged (Table~\ref{tab:s-split}). Without search, 0.4\% of its teams fall below the composition floor.

\begin{table}[h]
\centering
\caption{Splitting the root-held control's gain over player-aware greedy, 300 tournament lobbies, both seatings. Paired record (points per covered player, 95\% CI), J and population-judge consensus in \%.}
\label{tab:s-split}
\setlength{\tabcolsep}{3pt}
\begin{tabular}{llcccc}
\toprule
Arm & vs & Paired record & Cov. & J & Pop.\ judges \\
\midrule
Greedy's budget & player-aware greedy & +2.4 (0.6, 4.2) & 53\% & 55.5 & 52.4 \\
 & root-held & +0.3 ($-$1.1, 1.7) & 60\% & 49.5 & 49.3 \\
 & final drafter & $-$1.5 ($-$3.0, 0.1) & 57\% & 48.6 & 49.2 \\
 & players' own drafts & +3.3 (1.4, 5.2) & 56\% & 60.9 & 57.8 \\
 & population search & +4.0 (1.5, 6.6) & 36\% & 57.0 & 43.2 \\
Flat prior, greedy's candidates and bans & player-aware greedy & +1.3 ($-$0.4, 3.0) & 52\% & 49.4 & 49.7 \\
 & root-held & $-$3.6 ($-$5.5, $-$1.7) & 55\% & 44.4 & 47.6 \\
 & final drafter & $-$3.6 ($-$5.4, $-$1.7) & 52\% & 43.9 & 47.4 \\
 & players' own drafts & +0.9 ($-$0.9, 2.6) & 55\% & 56.1 & 56.3 \\
 & population search & +2.3 ($-$0.7, 5.3) & 35\% & 52.5 & 42.3 \\
\midrule
Greedy's bans & player-aware greedy & +0.9 ($-$0.8, 2.6) & 57\% & 54.6 & 52.7 \\
 & root-held & $-$0.9 ($-$2.3, 0.6) & 60\% & 48.5 & 50.8 \\
No assignment term & player-aware greedy & +1.3 ($-$0.3, 2.9) & 56\% & 54.3 & 53.1 \\
 & root-held & $-$0.7 ($-$2.2, 0.9) & 61\% & 48.7 & 51.1 \\
Greedy's set, own ranking & player-aware greedy & +2.3 (0.6, 4.1) & 54\% & 55.6 & 51.4 \\
 & root-held & +0.1 ($-$1.3, 1.5) & 60\% & 49.3 & 48.4 \\
\midrule
Prior only, no search & player-aware greedy & +2.2 (0.3, 4.0) & 55\% & 54.1 & 50.8 \\
 & root-held & $-$1.0 ($-$2.6, 0.5) & 58\% & 48.0 & 48.1 \\
 & final drafter & $-$0.4 ($-$2.0, 1.2) & 57\% & 47.8 & 47.7 \\
 & players' own drafts & +3.5 (1.6, 5.4) & 56\% & 60.7 & 57.2 \\
 & population search & +5.0 (2.3, 7.7) & 37\% & 57.2 & 42.6 \\
\bottomrule
\end{tabular}
\end{table}

\section{Banned best heroes}

This section takes the subset of the main text's natural experiment in which the best hero was banned rather than taken. When a player's best hero is banned, the player does 2.18 points worse (SE 0.19) in that game, against 1.38 predicted (1.19 million pick slots, 37,752 players). The comparison is observational, and bans may single out players in form, so we read the gap only as a hint that the model understates personal skill.

\section{Party study precision}

\begin{figure}[h]
\centering
\includegraphics[width=\columnwidth]{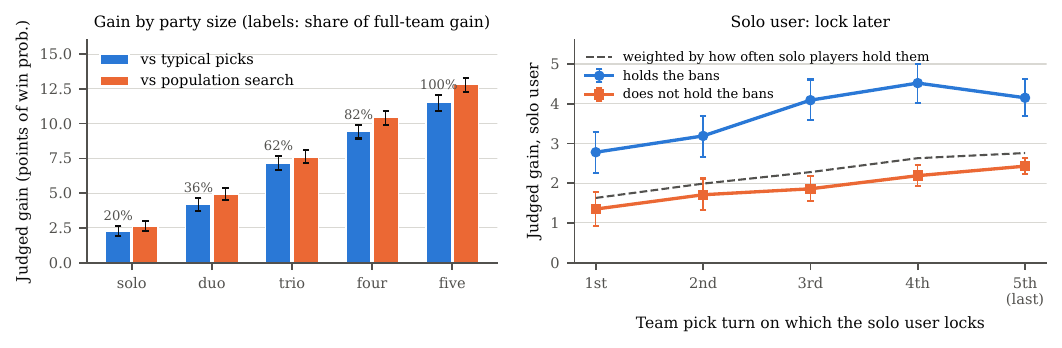}
\caption{Left: judged gain over typical picks (blue) and over population search on the same seats (orange) by party size; labels give the share of the full team's gain. Right: a solo player's gain by the team pick turn on which they lock, when they hold the bans and when they do not, and weighted by how often solo players hold them (19\%). 1,175 lobbies, 95\% CIs.}
\label{fig:s-party}
\end{figure}

With 1,175 lobbies the 95\% intervals for the solo lock turns are $\pm$0.18 to $\pm$0.37 points with both sides pooled. Paired gaps between turns are $\pm$0.2 to $\pm$0.4, tight enough to separate the early turns from the late ones but not the fourth from the last ($-$0.1 points, $-$0.4 to 0.1, weighted by how often solo players hold the bans).

\section{Matching lobby names to accounts}

When this runs live, the names visible in a lobby must be matched to accounts. In a simulation on 20,000 real lobbies, 73.8\% of players had an account in the public match records we collected. For those, the right account ranked first 93.9\% of the time, and accounts picked automatically were right 98.0\% of the time.

\section{Regions}

The players add 6.6 points in the Americas (57.4\% to 64.0\%) and 5.7 in Europe (57.1\% to 62.8\%) on the spring set, and 6.4 and 5.8 on the summer set. Korea (the game's Asia server region), 3 to 5\% of games, gains less: 5.1 points in spring and 3.7 in summer. With the base model on the two test sets together, the Korea gap is statistically clear ($-$1.5 points against the Americas, 95\% confidence interval $-$2.3 to $-$0.7, resampling games). Korean players recur often, so an interval that resamples players would be somewhat wider. Korean players have far shorter histories in our data: the median player in a Korean game has 58 earlier games, against 396 in the Americas and 315 in Europe. Korean accounts also have fewer recorded games in Heroes Profile itself: an active account (20 or more games in 2025) had a median of 101 games that year, against 160 to 172 elsewhere, so the thin histories come from upstream. At equal history depth Korea still gains about a point less, which we leave open.

\section{New accounts}

A new account can hide an experienced player. With the win-only model we flagged 1,327 such accounts, 10.2\% of new accounts with 20 or more games, using their results in their first 20 games, so their 32.6 points above expectation there reflect the flag itself. Outside the flagging games they stay ahead: 11.9 points above expectation in games 21 to 50 and 2.8 in games 51 to 100. Adjusting everyone else's records for them moved those records by 0.07 points on average and changed the headline accuracy by less than 0.01 points, because these games are a small share of any regular player's history. A separate starting estimate for genuinely new accounts, tested outside the full model, improved log loss by 2.4 thousandths, almost entirely in the 18\% of games that contain one. The full model has no such estimate: a new account starts where any player starts at zero games.

New accounts are those first seen on or after July 1, 2024 with low hero levels in their first ten games: 51,446 accounts, 13,048 with 20 or more games. An iterative two-group mixture over their first 20 games, each round correcting everyone else's records for flagged teammates and opponents, converged in 4 rounds and flagged 1,327 accounts. Flagged accounts appear in 6.3\% of games. A causal classifier at an account's 20th game separates the ones that stay hot with an AUC of 0.74, with scoreboard style as its best signal.

\section{Ground-truth worlds with players}
\label{sec:s-worlds}

\textit{Hero layer.} Each world starts from a draw of the hero-level outcome model of \pwof{}: per-hero strength with map and tier terms, pair synergy within a team, pair counters across teams, and a rank-8 term for three-hero interactions, fit to the training games and drawn from its approximate posterior, with independent effects for higher-order cells. The \emph{hero-only} worlds use this layer alone (its main family, with default dials), so players have no effect on outcomes.

\textit{Player layer.} The player worlds add a term for each player on each hero. In log-odds,
\begin{multline}
\operatorname{logit} P(\text{team 0 wins}) = z_{\text{hero}} \\
+ \sum_{(p,h) \in \text{team 0}} u'(p,h) - \sum_{(p,h) \in \text{team 1}} u'(p,h),
\end{multline}
with $u(p,h) = \gamma_{b(n_{ph})} + A_p + AP_{p,\tau} + R_{p,r(h)} + E_{ph}$ and $u'(p,h) = u(p,h) - \bar u_h$. Here $n_{ph}$ is the player's earlier ranked games on the hero, binned at 0, 1--2, 3--5, 6--10, 11--25, 26--50, 51--100 and 101 or more; $A_p$ is general skill beyond the rating; $AP_{p,\tau}$ is a player-by-half-year effect, so that skill can drift; $R_{p,r}$ is a player-by-role effect; and $E_{ph}$ is the player-hero cell. Centering by $\bar u_h$, the mean of $u$ over observed play, keeps the hero layer's averages intact. The layer is fit jointly with the population win probability as an offset, by maximum a posteriori with Gaussian priors, on 2,032,140 ranked games with all ten players identified, from January 2021 to June 30, 2026, before the judge window. The prior SDs are chosen by approximate evidence over a grid; the selected values are 0.05 logit for $A$, $AP$ and $E$ and 0.08 for $R$. Each world draws $A$, $AP$, $R$ and $E$ from their approximate posteriors, unseen players and cells from the prior, and a fresh half-year effect for the play period. The familiarity curve, relative to 101 or more earlier games, runs from $-$7.3 points for a hero never played to $-$1.4 points at 51 to 100 games. In sample, the player term lowers log loss from 0.6702 to 0.6324.

The layer is built differently from the drafters' player model on purpose. It has no borrowing across related heroes, no record pooling, no scoreboard and no recency weighting, so the drafters' estimates are imperfect measurements of a known truth, not copies of it.

\textit{f-out, f-cal and f-cov.} In f-out the player layer is fit on outcomes alone, as above. In f-cov it also includes the drafters' own skill estimate $s(p,h)$ and off-role flag as covariates, fit by logistic regression on half of the judge window (81,329 games, even game index) with the f-out term and the population win probability as an offset (coefficient 2.43, SE 0.07, on $s$; $-$0.04, SE 0.01, on the off-role flag), and centered by hero. f-cov is partly circular, because the drafters' estimates are part of the world's truth, and we label it so wherever it is used.

In f-cal the player term is again fit on outcomes alone, but from the same raw information the drafters use, in a different model class. A logistic regression with the population win probability as offset takes team differences of 156 features per slot: shrunk record and scoreboard summaries for the player on the hero, in the role and overall (all-time and with a 90-day half-life), log play counts and the familiarity bins. It uses no drafter state, estimate or calibration. It is fit on 322,332 games from January to June 2026, with features as of the first of each game's month, and the L2 penalty is chosen on June after fitting January to May. Each world adds residual player, player-period, player-role and player-hero draws with f-out's SDs, and scales the fitted term by a factor chosen, by a rule fixed before the check, as the smallest grid value whose expected slope reaches the real 0.74 (1.1; the unscaled fit is reported alongside). On the 162,657 held-out judge-window games, f-cal's log loss is 0.6570 (0.6565 unscaled), against 0.6652 for the drafters' own model, 0.6700 for f-out and 0.6789 for the population model alone. The scoreboard summaries close most of the gap to the real calibration; looser shrinkage or more play history alone reach only 0.3 to 0.5. Simply scaling f-out up to the real slope needs a factor of about 4 and makes held-out log loss worse than the population model alone (0.724), so it is not a plausible world.

\textit{Assignment.} Drafters choose heroes, not players. In each world a draft team's five players take its five heroes in the arrangement that maximizes the team's true player term (the best of 120). The main text reports this assignment. Random and human-like assignments, where an arrangement is drawn in proportion to $\prod (n_{ph} + 0.5)$, are computed alongside.

\textit{Checks.} The test is whether real outcomes follow the drafters' personal term as closely in the worlds as in real games. On the judge-window games with their real assignments (200 worlds, 162,657 games), we regress the residual on the drafters' personal term by the team's rarest-triple bin. In real games the slope is 0.74 to 0.84 across bins. In f-out it is 0.19 to 0.20 in every bin, below the real value in all 200 worlds: the outcome-only layer, fit without the drafters' estimates, ends up far less correlated with them than real outcomes are. In f-cov it is 0.76 to 0.77, with the real value inside the worlds' range in every bin (real values at the 23rd to 94th percentiles). In f-cal it is 0.68 to 0.73 (0.717, 0.688 to 0.747, at rarest triple 25 to 99, against a real 0.737). f-cal and f-cov therefore match the real data on this statistic and f-out does not, which is why the main text uses f-cal and f-cov.

\textit{Scoring.} Each arm drafts the 300 tournament lobbies against the same opponents as in the main matrix. Each world scores every final draft by its true win probability under the assignment above, and arms are compared on common opponents and paired over 1,000 worlds (mean and 95\% interval over worlds). In f-out, on the headline lobbies, the floor alone changes true value by +0.02 ($-$0.29 to 0.37) and the final drafter by +0.64 (0.37 to 0.94). The tune lobbies give the same comparisons: the floor alone changes true value by $-$0.38 ($-$1.02 to 0.24) in the hero-only worlds, $-$0.15 ($-$0.77 to 0.56) in f-cal, $-$0.14 ($-$0.77 to 0.59) in f-out and +0.06 ($-$0.51 to 0.80) in f-cov; the final drafter by +0.92 (0.54 to 1.30), $-$0.10 ($-$0.57 to 0.38), +0.71 (0.28 to 1.15) and +0.13 ($-$0.31 to 0.62). Under random rather than best assignment of players to heroes, the final drafter gains in f-cal (+0.67, 0.04 to 1.34, headline lobbies).

\textit{Truth-leaf regret.} Giving the final drafter's search the world's true value of every draft, with everything else fixed, gains 8.1 points against population search in the hero-only worlds, 6.7 in f-out and 3.6 in f-cov (against player-aware search with a player prior: 7.0, 6.0 and 5.2). The unconstrained selected design's regret is larger in every case but one (9.7, 7.5 and 4.0; and 8.8, 6.9 and 5.0). The floor's cost under the true value is 0.1 to 0.8 points. In the f-cal worlds (20 worlds, against population search) the final drafter's regret is 4.9 points (4.4 to 5.6) and the selected design's 4.6 (3.8 to 5.9); the floor's cost under the true value there is 0.2 ($-$0.2 to 1.1).

\section{Drift details}
\label{sec:s-drift}

\textit{The two clocks.} Table~\ref{tab:s-clocks} gives the values plotted in Fig.~\ref{fig:s-clocks}. Game-side tier groups are Bronze and Silver, Gold and Platinum, and Diamond and Master. A player who stops playing a hero has no value in window B, so players who stop cannot show a trivially unchanged value; player cells need two or more games in a window. Pooling gives 50 window pairs at 3 months and 29 at 2 years. Only 15\% of player-hero cells with two or more games in one window have two or more a year later. Year-old player records cost 17.68 thousandths of log loss through new players, 7.33 through newly played heroes and 9.48 through known pairs (34.50 summed, 33.49 for all slots at once, as the parts interact through the team difference); some of the new players existed but had no games in our corpus.

\begin{table}[h]
\centering
\caption{Correlation of true values between two 90-day windows, by the gap from the end of one to the start of the other (95\% bootstrap CI). Game: a hero's win rate in a tier group. Player: skill on a hero relative to the player's own level, over cells with 2+ games in each window (all cells) or 20+. The noise correction is uncapped, so an interval can pass 1.}
\label{tab:s-clocks}
\setlength{\tabcolsep}{3pt}
\begin{tabular}{lccc}
\toprule
Gap & Game & Player, all cells & Player, 20+ games \\
\midrule
3 months & 0.93 (0.90, 0.95) & 0.91 (0.80, 1.04) & 0.83 (0.76, 0.93) \\
6 months & 0.90 (0.87, 0.93) & 0.85 (0.75, 0.98) & 0.79 (0.68, 0.92) \\
1 year & 0.85 (0.79, 0.89) & 0.90 (0.78, 1.08) & 0.74 (0.61, 0.90) \\
2 years & 0.77 (0.69, 0.83) & 0.86 (0.67, 1.11) & 0.57 (0.38, 0.84) \\
\bottomrule
\end{tabular}
\end{table}

\textit{Reworked heroes.} Correlation of a hero's tier win rate between windows, by the number of patches between the windows that list the hero with a balance or rework change: at 1 year, 0.91 with no listed change, 0.86 changed once and 0.79 changed twice or more.

\textit{Regions.} Hero win-rate correlation at 3 months, 6 months and 1 year: Americas 0.93, 0.85, 0.73; Europe 0.94, 0.87, 0.76; Korea 0.88, 0.84, 0.77.

\textit{Deployment replay.} Regret against retraining after every build, over 2.8 years, as the mean $\pm$ standard deviation over three training seeds (42, 123, 777), each scored against its own retrain-every-build reference (56.87\% to 56.89\% accuracy): never retrain or refresh, 1.00 $\pm$ 0.03 points; refresh every build without retraining, 0.64 $\pm$ 0.03; retrain every 3 builds, 0.02 $\pm$ 0.02 with refresh between and 0.05 $\pm$ 0.02 without; every 6 builds, 0.06 $\pm$ 0.03 with refresh and 0.12 $\pm$ 0.02 without; every 9 or 12 with refresh, 0.17 $\pm$ 0.02 and 0.17 $\pm$ 0.03. The ordering of the policies, and the gain from refreshing between retrains, hold in every seed.